\documentclass[fleqn,10pt]{wlscirep}
\usepackage[utf8]{inputenc}
\usepackage[T1]{fontenc}
\usepackage{lineno}
\usepackage{times}
\usepackage{epsfig}
\usepackage{graphicx}
\usepackage{caption}
\usepackage{subcaption}
\usepackage{amsmath}
\usepackage{amssymb}
\usepackage{mathtools}
\usepackage{caption}
\usepackage{subcaption}
\usepackage{cleveref}
\usepackage[most]{tcolorbox}
\usepackage{wrapfig}

\usepackage{lineno}
\usepackage{hyperref}
\usepackage{url}
\usepackage{lscape}
\usepackage{booktabs}
\usepackage{amsmath}
\usepackage{rotating}
\usepackage{multirow}
\usepackage{graphics}
\usepackage{rotating}
\usepackage{verbatim}
\usepackage{float} 

\title{Enhancing Food Intake Tracking in Long-Term Care with Automated Food Imaging and Nutrient Intake Tracking (AFINI-T) Technology}

\author[1,2,3*+]{Kaylen~J.~Pfisterer}
\author[3,4+]{Robert~Amelard}
\author[1,3,4]{Jennifer~Boger}
\author[1,2]{Audrey~G.~Chung}
\author[3,5]{Heather~H.~Keller}
\author[1,2,3]{Alexander~Wong}
\affil[1]{University of Waterloo, Waterloo, Systems Design Engineering,  Waterloo, ON, N2L 3G1, Canada}
\affil[2]{Waterloo AI Institute, Waterloo, ON, N2L 3G1, Canada}
\affil[3]{Schlegel-UW Research Institute for Aging, Waterloo, N2J 0E2, Canada}
\affil[4]{KITE-Toronto Rehabilitation Institute, University Health Network, Toronto, ON M5G 2A2}
\affil[5]{University of Waterloo, Waterloo, Kinesiology and Health Sciences, Waterloo, ON, N2L 3G1, Canada}

\affil[*]{kpfisterer@uwaterloo.ca}

\affil[+]{these authors contributed equally to this work}


\begin{abstract}

Half of long-term care (LTC) residents are malnourished increasing hospitalization, mortality, morbidity, with lower quality of life. Current tracking methods are subjective and time consuming. This paper presents the automated food imaging and nutrient intake tracking (AFINI-T) technology designed for LTC. We propose a novel convolutional autoencoder for food classification, trained on an augmented UNIMIB2016 dataset and tested on our simulated LTC food intake dataset (12 meal scenarios; up to 15 classes each; top-1 classification accuracy: 88.9\%; mean intake error: -0.4 mL$\pm$36.7 mL). Nutrient intake estimation by volume was strongly linearly correlated with nutrient estimates from mass ($r^2$ 0.92 to 0.99) with good agreement between methods ($\sigma$= \textminus2.7 to \textminus0.01; zero within each of the limits of agreement). The AFINI-T approach is a deep-learning powered computational nutrient sensing system that may provide a novel means for more accurately and objectively tracking LTC resident food intake to support and prevent malnutrition tracking strategies.

\end{abstract}
\begin{document}

\flushbottom
\maketitle

\thispagestyle{empty}

\section{Introduction}
Malnutrition leads to higher morbidity~\cite{pirlich2001}, and lower quality of life~\cite{keller2004}. In the United States, malnutrition imparts over 4 times higher odds of hospitalization and \$21,892 more in total charges per stay~\cite{lanctin2021prevalence}. It is clear that nutritional status has multidomain effects with both fiscal and clinical ramifications and should be monitored. Older adults (aged 65 years plus) living in long-term care (LTC) homes are especially nutritionally vulnerable in part due to low food intake~\cite{keller2017prevalence}. More specifically, in Canada, 54\% of LTC residents are either malnourished or at risk for malnutrition~\cite{keller2019prevalence}. This is higher than global estimates ranging from 19\% to 42\% (37 studies, 17 countries)~\cite{bell2013prevalence}. Additional independent risk factors for malnutrition are eating challenges, and increased cognitive impairment~\cite{keller2017prevalence, vucea2017} which describes between 47\% to 90\% of the Ontario LTC population~\cite{CDC2013,CIHI_2018}. Thus, tracking and preventing poor food intake is essential for supporting healthy aging. 

However, there is a lack of objective and quantitative tracking methods for food and fluid intake especially for centralized intake tracking by proxy (i.e., multiple staff tracking a set of residents' intake). Registered dietitian (RD) referrals get triggered and nutritional support system effectiveness is monitored based on nutritional assessment best practices including: unintentional weight loss, usual low intake of food~\cite{DoC_2019}.  Resident food and fluid intake charting completed by either personal support workers (PSWs) or nursing assistants capture intake across a meal via visual assessment within 25\% incremental proportions at the end of the meal but may be completed hours after due multiple competing priorities during mealtime. As a result, due to inconsistency and subjectivity in charting methods, about half of residents who would benefit from an intervention are missed~\cite{simmons2000nutritional,simmons2006feeding}. 
Furthermore, there is a lack of trust in current methods because they are known to have poor accuracy and validity~\cite{martin2008,williamson2003} limiting clinical utility. However, it raises awareness to some extent regardless of whether the measurements are inaccurate (e.g., food spills). Measuring food intake is a proxy for nutritional status, however it provides a sense of \textit{why} something might be going wrong (in combination with biomarkers). Better, more reliable measurements would enable more meaningful assessment of probing when, how, and why, something may be going wrong to better inform intervention strategies and care providers have expressed a desire to leverage higher quality data, provided they are reliable and trustworthy~\cite{pfisterer2019}. 

Automated tools may provide a palatable solution which removes subjectivity and with higher accuracy than human assessors. This may also enable time efficient measurement of food intake at the energy, macro and micronutrient levels~\cite{pfisterer2019}. To estimate food intake and nutrient consumption, four main questions must be answered: \textit{where} is there food (segmentation); \textit{which} foods are present (classification);  \textit{how much} food was consumed (pre-, post-prandial volume estimation); and \textit{what} was the estimated food and nutrient intake? We have previously investigated \textit{where} food is and  {how much} food was consumed at a bulk intake level~\cite{pfisterer2021segmentation}. For the purpose of this paper, we focus on \textit{which} foods are present, \textit{how much} food was consumed for enabling assessment of \textit{what} was the estimated food intake at the nutrient level. Our proposed automated food imaging and nutrient intake tracking (AFINI-T) technology measures food intake compared against ground truth weighed food records, addresses automatic segmentation with integrated RGB-D assessments, and was evaluated in both regular and modified texture foods. 

The purpose of this research was to explore nutritional composition estimation using  AFINI-T with an integrated RGB-D camera for co-aligned colour and depth image acquisition, and to quantify the error of computing nutritional intake from relative changes in volume as compared to ground truth nutritional intake from gold standard weighed-food records. Here, we highlight additional requirements identified through interviews with end-users and provide an assessment of the proposed system around this context. The remainder of this paper is organized as follows: an overview of related work; results from this case study including user needs translated to design requirements; experimental results including segmentation, classification, volume estimation, bulk intake, and nutrient intake accuracies; a discussion including current limitations, and future directions; and finally, an overview of methods pertaining to the case study and experimental procedures for technical implementation.

\subsection{Related work}
While automated tools may provide a time efficient, and objective alternative, little work has been done within the LTC context. Since LTC residences are at high risk for malnutrition, tracking food intake is important and there are large demands on staff time, an automated system has potential to provide high impact in this field. 

Certainly there has been progress made in food classification, however, these previous efforts tended to focus on classification in isolation (e.g.,~\cite{lo2020image,bruno2017survey,doulah2019systematic,pouladzadeh2016,subhi2019}). Food segmentation is comparatively unexplored but has typically required fiducial markers~\cite{okamoto2016,zhu2015}, multiple images~\cite{kong2015dietcam,pouladzadeh2014measuring}, manual labelling for \textit{every} food item and each food image~\cite{meyers2015}, or did not predict food areas at the pixel-level (e.g., bounding boxes\cite{shimoda2015cnn,kawano2015}). In a LTC setting, semantic segmentation is required when measuring food and nutrient \textit{intake} across a plate as many residents do not consume their entire portion. For food volume estimation, template matching has been popular~\cite{he2013food,xu2013,chae2011,jia2014accuracy,ofei2019validation,rachakonda2020ilog,herzig2020volumetric}. However, within the context of LTC, the same food can take on various shapes (e.g., banana in peel versus sliced banana, versus pureed banana) and 47\% of the LTC population receives modified texture foods~\cite{vucea2019prevalence} which limits utility of template-matching in this context. Others have leveraged stereo reconstruction~\cite{dehais2017,puri2009recognition} for volume estimation and 3D point-clouds~\cite{rahman2012food}. However, taking multiple images and building 3D point clouds require additional time and precision, an extremely limited commodity in the LTC context. Staff in LTC require something quick and simple and solutions that take time, effort, and precision are both difficult to implement while running the risk of not being done consistently.

An alternative route that has been gaining popularity is depth imaging and structured lighting for mapping food topology~\cite{fang2016,shang2011,meyers2015,chen2012,liao2016}. While depth-only and structured-light-only approaches are more robust to illumination variation, highly reflective foods (e.g., gelatin, soup) pose a challenge for accurate readings~\cite{liao2016}. Leveraging structured light for measuring volume shows promise as a means to address previous shortcomings particularly with the respect to required operator time with accuracy approaching 97.44\%~\cite{liao2016}. While structured light has been gaining popularity in the agrifood industry for measuring volume~\cite{long2018}, monitoring fermentation in bread~\cite{ivorra2014,verdu2015}, classifying fruits and vegetables~\cite{perez2017} and apple quality analysis~\cite{lu2017,lu2018}, there has been little work on incorporating structured lighting for monitoring food intake~\cite{shang2011,shang2012,liao2016,fang2016}. Shang \textit{et al.} imaged only 9 food replicas and did not report error~\cite{shang2011,shang2012}, and Fang \textit{et al.} reported error of 2\% but did not disclose how many foods or what was used as ground truth~\cite{fang2016}. Consistent with~\cite{fang2016}, Liao \textit{et al.} reported an average error in weight of 7.5\%~\cite{liao2016} both showing a marked improvement over human assessors method fraught with incorrect measurements up to 66\% of the time when assessed retrospectively~\cite{castellanos2002}. One potential reason for few papers leveraging a structured light approach is that until recently, these systems were difficult or expensive to build. However, with recent advances in commodity RGB-D cameras such as the Microsoft Kinect and Intel RealSense has enabled the incorporation of depth analysis at relatively low cost (i.e., under \$200) or comes standard in newer versions of iPhones and iPads.

An additional factor limiting progress in food \textit{intake} estimation more broadly is the need for large, complete food datasets~\cite{zhou2019application}. Especially when considering food \textit{intake} estimation in LTC, ground truth nutritional values and pixel-wise labelling is crucial in addition to images collected from a consistent perspective, having good colour representation, good inclusion of modified texture foods, capturing complex meal scenarios (e.g., multiple foods on a plate, prepared foods), and having open-source access to the data~\cite{pfisterer2021segmentation}. Based on a survey, we are aware of only one dataset, UNIMIB2016~\cite{cioccaJBHI} that addressed the majority (though not all) of those requirements (multiple prepared foods, consistent imaging, pixel-wise annotation) which is why we systematically curated our own regular and modified texture simulated food intake dataset as described in~\cite{pfisterer2021segmentation}. As a result, few attempts have been made to solve this problem, they have high error (up to 400~mL error~\cite{meyers2015}), or are limited to synthetic foods~\cite{lo2018food,lo2019point2volume} which do not accurately represent a real-world environment. While we can borrow inspiration from these approaches, they were not designed with the needs of LTC in mind. For example, weight loss was the desired outcome~\cite{kong2015dietcam,meyers2015im2calories,okamoto2016,pouladzadeh2016, aslan2018semantic,aguilar2018grab} which is inappropriate in the LTC context and they were designed to be used by the user. In LTC we have proxy users where one staff monitors many residents, thus there is a unique opportunity for novel methods to support the requirements within the LTC setting to support food and fluid intake tracking best practices. 

There has been promising work on technology-driven methodologies broadly as summarized in~\cite{doulah2019systematic,boushey2017,selamat2020automatic,moguel2019systematic,vu2017wearable}. Many have taken a wearable sensor approach typically developed for individual use (in contrast to agents monitoring intake of others). These wearable sensor approaches typically include video cameras, microphones or strain sensors. While promising, within the LTC domain, a wearables approach is infeasible when considering the need for privacy, eating assistance and the requirement for frequent sterilization. Additionally, budget constraints limit the feasibility when multiple devices would be required to monitor intake of residents. There is also the consideration of how a wearable device may impact food intake. For example, while micro-cameras can improve intake estimates, feelings of self-consciousness may change eating behaviour towards lower intake as has been seen in both adults and children~\cite{pettitt2016pilot,beltran2016adapting}. LTC requires a different kind of cost-effective solution capable of time-efficient and large-scale monitoring. As food is plated from a centralized location (e.g., in a kitchenette), an image-based approach conducted in a semi-permanent spot is a viable option that could overcome many of these challenges.

An additional consideration is the need to disentangle sources of error as error assessment is typically not reported or segmentation is coupled with either classification~\cite{pouladzadeh2016food,kawano2015,he2013food,meyers2015,shimoda2015cnn,wang2018,yunus2018framework} or volume estimation~\cite{hassannejad2017}. Others have also identified this as a limitation for accurately predicting down-stream nutritional information~\cite{meyers2015,aslan2018semantic,wang2018,zhu2015}. These gaps have been corroborated by recent reviews identifying the need for automated systems~\cite{subhi2019,doulah2019systematic} that user more representative foods~\cite{pouladzadeh2016,subhi2019} with the need for more in-depth evaluation and statistical analyses~\cite{doulah2019systematic}. This work seeks to address this gap in the context of LTC. 

Technologies for food intake tracking in LTC has been understudied. What has been done has either involved assessment of food waste using visual estimation and digital photography involving significant operator time~\cite{parent2012}, or an electronic nutrient tracking system assessed for agreement with food diaries (i.e., the current accepted method)~\cite{astell2014validation}. We seek to go beyond current methods for improving intake estimation accuracy using gold standard weighed food method as ground truth for validation. Borrowing from the hospital setting, a more recent pilot study used a combination of RFID scanner, built-in food scale and digital photography method for estimating portion size across a plate at a time~\cite{ofei2019validation}. They yielded promising results in terms of accuracy with relatively high agreement between trained and untrained assessors (ICC = 0.88, \textit{P}$<$0.01) and excellent agreement for protein and energy intake between their method and the weighed food method (ICC =0.99, \textit{P}$<$0.01)~\cite{ofei2019validation}. These are encouraging results for an adjacent application in LTC. Based on our previous work, however, in LTC it is desirable to have a higher level of detail including an intake breakdown for each item consumed (not averaged across a plate)~\cite{pfisterer2019}. This paper describes the final stage of pixel-wise food classification and nutrient linking through intake prediction for providing food and nutrient intake estimation specifically designed for LTC. It builds on previous work~\cite{pfisterer2021segmentation} leveraging a specialized food-segmentation method powered by deep learning for automated segmentation, moving from bulk food segmentation to nutritional estimation with a few additional steps modularized for systematic error assessment.

\section*{Results}
This research focused on the characterization of changes in volume on a whole plate level for bulk intake estimation reporting degree of consumption (i.e., proportion of food consumed) as well as nutritional intake estimation using a nutritional look-up table at the food-item and whole plate level. When considering the application realities and needs surrounding use in LTC, and specifically how classification might be applied, an additional point for consideration is limited computational resources. To meet the needs of this environment, we require (a) a salient feature extractor that can be trained in advance and supports real-time use, (b) a classification method that is light-weight for mobile application use, and (c) a daily, easily updatable classifier to account for \textit{a priori} menu plans. The results presented here include qualitative results which have been translated into design requirements (Section~\ref{Results - Case Study}), followed by quantitative experimental results for segmentation, volume estimation, classification, and nutrient intake (Section~\ref{experimental_results}).

\subsection{Design requirements: A case study for motivating focus areas}
\label{Results - Case Study}

Data from interviews and a workshop with target end-users (e.g., primary users: nurses, personal support workers, RD; secondary users: directors/assistant directors of food services, dietary aids, neighbourhood coordinators, recreation team) as part of implementing participatory-iterative design  provide additional insights beyond what was previously from the user study work described in~\cite{pfisterer2019}. These results were translated into design requirements to inform technical implementation, as described below.

\subsubsection{Role of the registered dietitian and appropriate referrals}

Regarding the interviews, ultimately it was clear that the RD needs to be the gate-keeper to ensure any changes to nutritional management is holistically assessed. The interview with a LTC RD conveyed that, in Ontario, RDs get one half-hour block per resident each month. This isn't much time. The ablity to streamline the workload by automating manual tasks (e.g., manual calculations of individualised targets) may help to offset the time required and allow the RD to assess resident nutritional status more deeply. This RD saw the most potential for this technology to help make more appropriate referrals and to decrease the data overload of multiple, unranked referrals. 

\subsubsection{Need for fine-grained tracking and nutrients of interest}
The LTC RD indicated that more accurate intake tracking would help especially considering there is no current discrimination between which foods were consumed. For example, if 25\% of the main dish was consumed, there is no way to know if it was the protein or the vegetable; it assumes an even distribution across the foods offered on the plate. For this reason, the AFINI-T system must provide more precise information pertaining to the types of foods consumed, and track at a lower level (nutrient level) so key nutrients with clinical endpoints can be better monitored. This RD identified the following clinical endpoints: falls (calcium, vitamin D, protein), new or worsening pressure ulcers (protein), constipation (fibre, fluids), and weight loss (calories, protein). Similarly, zinc, vitamin K, and vitamin B6 are often poorly consumed in this population with clinical relevance pertaining to cognition, wound healing and immunity~\cite{huskisson2007,reginaldo2014,keller2018prevalence}. We took the union of nutrients of interest identified here along with the nutrients of interest identified through the design process (calories, carbohydrates, protein, fats, calcium, iron, sodium and vitamin C). This resulted in a final set of 13 nutrients of interest. \textbf{Macronutrients}: calories, carbohydrates, fats, fibre, and protein. \textbf{Micronutrients}: calcium, iron, sodium, vitamin B6, vitamin C, vitamin D, vitamin K, and zinc.

\subsubsection{The current workflow}

The interview with the RD nutrition research expert indicated that currently food intake is tracked by team members walking around the dining room implying any sort of food imaging will impart an increased time. Furthermore, in LTC the emphasis is more on quality of life and honouring resident preferences, additionally corroborated during the quarterly registered dietitian meeting. Typically, even menu planning is conducted at the macronutrient level, so this type of system has potential to meet resistance regardless of the analytical power without some form of incentivisation. Based on workshop discussion including perspectives of dietary aides, recreation staff, and directors and assistant directors of food services, they expressed the desire to more precisely record food and fluid intake to inform individual interventions and menu-planning more broadly. It seems there is a strong desire to provide enhanced care and make better use of tracking information than what is mandated, but that barrier to do so is the inaccuracy of the current system. This suggests that data-driven insights may be perceived as having higher utility providing incentives inherently. 

\subsubsection{Real-time use}

Regarding practicalities of real-time use, insight from the workshop motivated this work. One main emerging theme was the need for the system to work even when internet is down. Users need to log into the current system to track nutrition. If it was down, then users could not log anything and often required retrospective charting in these situations. There was enthusiasm for compatibility offline and users would prefer the option to save offline with the ability to sync when able both for when the internet is down or to support dining field trips to outside restaurants. For more than just privacy concerns and considerations, this further corroborates the need for an on-board system. This applies both to the computational requirements of the system as well as how classification models are updated. 

\subsubsection{Leveraging \textit{a priori} information}
Instead of treating classification in isolation, in LTC there are several insights driving how food is prepared and served that can be used to limit the set of menu item options. First, menus are planned in advance. This provides crucial information in terms of limiting possible foods as well as providing an opportunity to leverage menus planned with Canada's Food Guide~\cite{CFG2019} in mind for the purpose of providing a nutrient level baseline of what is in each portion of food. Second, certain foods are more likely to be served at certain times of the day. While there may be exceptions to this rule for honouring preferences, from a constraints perspective this enables us to limit the potential possibilities and simplify the classification problem. Instead of needing to ascertain if a sandwich was from any number of restaurants, we can assume it is most likely from the menu for a given day and again more likely for that sandwich to be served at lunch. With this approach to constraining the search space, it becomes feasible to build a system that achieves high accuracy within a naturally constrained environment. One foreseen challenge with the modified texture foods is discrimination between the same food at different levels of modification. While they are the same foods, they are prepared differently (e.g., broth is added to pur\' ees) making it important to properly discriminate between sub-categories of the same food. Here, we make use of the additional prior of texture to further refine based on the level of texture prescribed. For example, if a resident receives pur\' eed foods, the regular and minced options for that meal would be automatically excluded from possibilities, simplifying the number of potential food classes for that resident's meal. 

\subsection{Experimental results}
\label{experimental_results}
These quantitative results provide an overview of the AFINI-T systems food and nutrition intake estimation system. Here we provide details on the AFINI-T system's strategies for segmentation, classification, volume estimation, bulk intake, and nutrient intake accuracies.

\subsubsection{Segmentation accuracy}
\label{resultsseg}
Table~\ref{tab:avg_seg_class} provides an overview of segmentation accuracy. Generally, results represent two types of meal scenarios: multiple regular texture foods (RTF) dataset on one plate , and single modified texture foods (MTF) dataset on a plate. The RTF dataset had 9 unique foods across 375 simulated intake plates. The MTF dataset foods were prepared by the LTC kitchen and included 93 unique foods including both pur\' ees and minced foods across 665 simulated intake plates. Across the RTF and MTF datasets there are 104 classes represented in 1040 simulated intake plate images. Segmentation accuracy was good with an average an intersection over union (IOU) of 0.879 across RTF and MTF datasets (Table~\ref{tab:avg_seg_class}). Segmentation accuracy ranged from 0.823 on the MTF dataset at lunch, to 0.944 on the RTF dataset for breakfast. From the perspective of IOU, the MTF dataset was more poorly segmented by the EDFN-D, however, consistent with~\cite{pfisterer2021segmentation}, the degree of visual-volume discordance is higher for modified texture diets and is discussed further in Section~\ref{resultsvol}. 

\begin{table}[H]
\centering
\caption[Average segmentation and classification accuracies within and across datasets.]{Average segmentation and classification accuracies within and across datasets.}
\label{tab:avg_seg_class}
\begin{tabular}{llcc}
\hline
\multicolumn{2}{l}{\textbf{Dataset}} & \textbf{Segmentation accuracy} & \textbf{Classification Accuracy} \\
\textbf{Meal} & \textbf{\begin{tabular}[c]{@{}l@{}}\# of classes\\ (\# images)\end{tabular}} & \textbf{IOU} & \textbf{Top-1} \\ \hline
RTF: Breakfast & 3 (125) & 0.944 $\pm$ 0.019 & 93.5\% \\
RTF: Lunch & 3 (125) & 0.919 $\pm$ 0.033 & 93.5\% \\
RTF: Dinner & 3 (125) & 0.928 $\pm$ 0.019 & 95.1\% \\
\textit{        RTF subtotal} & \textit{9 (375)} & \textit{0.929 $\pm$ 0.027} & \textit{93.9\%} \\ \hline
MTF: Day 1 - Lunch & 5 (25) & 0.841 $\pm$ 0.123 & 89.0\% \\
MTF: Day 1 - Dinner & 15 (90) & 0.823 $\pm$ 0.099 & 70.2\% \\
MTF: Day 2 - Lunch & 12 (74) & 0.863 $\pm$ 0.118 & 70.6\% \\
MTF: Day 2 - Dinner & 12 (91) & 0.840 $\pm$ 0.122 & 64.9\% \\
MTF: Day 3 - Lunch & 10 (85) & 0.834 $\pm$ 0.132 & 80.4\% \\
MTF: Day 3 - Dinner & 15 (109) & 0.859 $\pm$ 0.100 & 70.4\% \\
MTF: Day 4 - Lunch & 9 (60) & 0.871 $\pm$ 0.113 & 72.2\% \\
MTF: Day 4 - Dinner & 10 (90) & 0.837 $\pm$ 0.107 & 67.8\% \\
MTF: Day 5 - Lunch & 5 (41) & 0.881 $\pm$ 0.117 & 87.8\% \\
\textit{        MTF subtotal} & \textit{93 (665)} & \textit{0.849 $\pm$ 0.116} & \textit{73.7\%} \\ \hline 
\textbf{TOTAL} & \textbf{104 (1040)} &\textbf{ 0.879 $\pm$ 0.101} & \textbf{88.9\%} \\ \hline
\multicolumn{4}{l}{\begin{minipage}[t]{0.8\textwidth}There were no samples for Day 5 Dinner. RTF: Regular texture foods; MTF: Modified texture foods. See \textit{\textbf{Supplementary Table }S1 }for a full list of included foods.\end{minipage}}

\end{tabular}
\end{table}

\subsubsection{Classification accuracy}
\label{resultsclass}
As shown in Table~\ref{tab:avg_seg_class}, classification accuracy was higher for the regular texture dataset with top-1 accuracy (i.e., the most likely class) ranging from 93.5\% on breakfast and lunch to 95.1\% on dinner. However, the regular texture dataset had only three classes per meal, so it was a less challenging classification problem compared to a greater number of classes to differentiate between, especially when considering the modified texture dataset had less texture variance. In contrast, the modified texture dataset ranged from 64.9\% on Day 2 Dinner with 12 classes, to 89.0\% on Day 1 Lunch with 15 classes.

\subsubsection{Volume estimation accuracy} \label{resultsvol}
\begin{figure}[H]
	\centering
	\includegraphics[width=\textwidth]{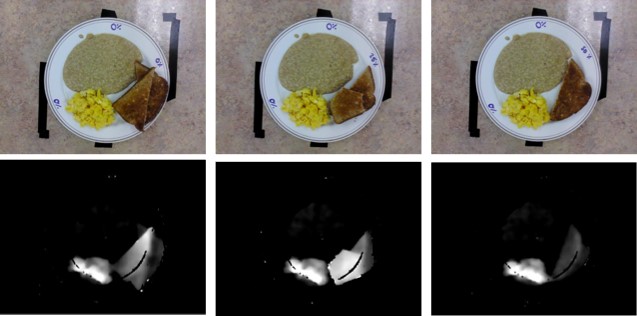}
	\caption[The toast occlusion conundrum.]{The occlusion conundrum, as demonstrated by stacked toast with an overhang. Since volumetric food estimation is based on pixel-wise classification, the pixels of the overhang are assumed to contain toast as a limitation to overhead imaging and also translates to having a pocket of air beneath providing a simplified example of low-density foods as well (e.g., salad). In addition, a similar issue is observed if it is placed on an angle up the side of the plate since toast is rigid. This is seen in the depth images where the brighter the pixel is the closer it is to the camera; brighter pixels are foods of greater height. Toast is a rigid plane and in the first example, we see a gradient from lower to higher near the tip with a similar, but less obvious trend in the third depth image. The depth map range was adjusted to exemplify the toast height.}
    \label{fig:toast_conundrum}
\end{figure}

Low density foods pose challenges for depth scanning systems. Here, volume estimation was within tolerance with food volume error of 2.5 $\pm$ 9.2 mL, and low-density foods (e.g., salad) have the largest food volume error seen for RTF:L of -10.1 $\pm$ 22.2 mL. A similar issue of low-density foods is seen through the 3D \% absolute error intake of 14.4 $\pm$ 13.1 \% which we suspect is due to the air pocket below some of the pieces of toast which sit on a tangential angle to the plate, or when two pieces are stacked with overhang as shown in Figure~\ref{fig:toast_conundrum}. This could be considered one of the classic examples of the ``occlusion conundrum'' with the imaging limitation of collection from an overhead view. This is an example of where segmentation could be done perfectly but would translate to volume estimation errors.

\subsubsection{Bulk intake accuracy}
\label{resultsbulkintake}

Table~\ref{tab:bulk_intake_accuracy} summarizes the bulk intake accuracy within and across datasets. Compared to~\cite{pfisterer2021segmentation}, for this iteration, we incorporated more representation of green in the UNIMIB+ dataset for training and validation, and introduced a more optimal stop-criteria for training for segmentation. In~\cite{pfisterer2021segmentation}, we saw the mean absolute volume error was 18.0$\pm$50.0 mL for regular texture foods (RTFs) and 2.3$\pm$3.2 mL for modified texture foods (MTFs) and mean volume intake error of 130.2$\pm$154.8 mL and 0.8$\pm$3.6 mL for RTFs and MTFs, respectively. Here, accuracy is higher with mean absolute food volume error of 6.6$\pm$13.6 mL for RTFs, and 2.1$\pm$3.1 for MTFs. Similarly, the bulk intake accuracy is higher with mean absolute intake error is greatly reduced for the RTFs with 39.9$\pm$39.9 mL but slightly higher for MTFs 6.0$\pm$5.6 mL. The higher degree of visual-volume discordance for MTFs compared to RTFs is again corroborated in Table~\ref{tab:bulk_intake_accuracy} with the mean food volume error of 3.8 mL $\pm$ 8.8, and higher mean volume error on the RTF dataset (6.6$\pm$13.6 mL) than for the MTF dataset (2.1$\pm$3.1 mL).


\begin{sidewaystable}[]
\caption[Bulk intake accuracy within and across datasets.]{Bulk intake accuracy within and across datasets.}
\label{tab:bulk_intake_accuracy}
\centering
\begin{tabular}{llcccccc}
\hline
\multicolumn{2}{c}{\textbf{Dataset}} & \multicolumn{2}{c}{\textbf{Food volume error}} & \multicolumn{4}{c}{\textbf{Bulk intake accuracy}} \\
\textbf{Meal} & \textbf{\begin{tabular}[c]{@{}l@{}}\# of classes\\ (\# images)\end{tabular}} & \textbf{\begin{tabular}[c]{@{}l@{}}Mean absolute error,\\ food volume (mL)\end{tabular}} & \textbf{\begin{tabular}[c]{@{}l@{}}Food volume \\ error (mL)\end{tabular}} & \textbf{\begin{tabular}[c]{@{}l@{}}Mean absolute\\ error, intake (mL)\end{tabular}} & \textbf{\begin{tabular}[c]{@{}l@{}}Error,\\ intake (mL)\end{tabular}} & \textbf{\begin{tabular}[c]{@{}l@{}}3D \% absolute\\  intake error\end{tabular}} & \textbf{3D \% intake error} \\ \hline
RTF: B & 3 (125) & 3.0 $\pm$ 4.1 & -2.2 $\pm$ 4.5 & 17.0 $\pm$ 14.3 & -15.1 $\pm$ 16.3 & 14.4 $\pm$ 13.1 & -12.0 $\pm$ 15.3 \\
RTF: L & 3 (125) & 11.0 $\pm$ 21.7 & -10.1 $\pm$ 22.2 & 76.1 $\pm$ 48.5 & 18.1 $\pm$ 88.7 & 13.7 $\pm$ 9.0 & 7.6 $\pm$ 14.6 \\
RTF: D & 3 (125) & 6.0 $\pm$ 6.0 & -5.8 $\pm$ 6.1 & 26.5 $\pm$ 14.4 & -24.5 $\pm$ 17.7 & 11.2 $\pm$ 9.9 & -2.9 $\pm$ 14.7 \\ \hline
\textit{        RTF subtotal} & \textit{9 (375)} & \textit{6.6 $\pm$ 13.6} & \textit{-6.0 $\pm$ 13.9} & \textit{39.9 $\pm$ 39.9} & \textit{-7.2 $\pm$ 56.0} & \textit{13.1 $\pm$ 10.9} & \textit{-2.5 $\pm$ 16.8} \\ \hline
MTF: D1 - L & 5 (25) & 1.0 $\pm$ 1.1 & -0.7 $\pm$ 1.3 & 3.4 $\pm$ 3.3 & -0.9 $\pm$ 4.7 & 5.0 $\pm$ 4.6 & -0.3 $\pm$ 6.9 \\
MTF: D1 - D & 15 (90) & 1.9 $\pm$ 2.9 & -1.1 $\pm$ 3.3 & 4.1 $\pm$ 3.7 & 2.5 $\pm$ 5.0 & 7.4 $\pm$ 14.1 & 6.4 $\pm$ 14.6 \\
MTF: D2- L & 12 (74) & 2.2 $\pm$ 3.3 & 0.0 $\pm$ 4.0 & 7.4 $\pm$ 7.3 & 6.1 $\pm$ 8.4 & 6.7 $\pm$ 5.5 & 5.1 $\pm$ 7.0 \\
MTF: D2 - D & 12 (91) & 1.2 $\pm$ 1.0 & 0.1 $\pm$ 1.5 & 4.6 $\pm$ 4.3 & 2.9 $\pm$ 5.5 & 8.3 $\pm$ 7.7 & 5.5 $\pm$ 9.9 \\
MTF: D3 - L & 10 (85) & 3.8 $\pm$ 5.1 & -3.3 $\pm$ 5.5 & 7.6 $\pm$ 6.3 & 5.0 $\pm$ 8.5 & 11.5 $\pm$ 10.0 & 10.0 $\pm$ 11.5 \\
MTF: D3 - D & 15 (109) & 1.9 $\pm$ 2.0 & 0.3 $\pm$ 2.7 & 5.5 $\pm$ 3.8 & 3.9 $\pm$ 5.4 & 6.7 $\pm$ 4.7 & 4.9 $\pm$ 6.6 \\
MTF: D4 - L & 9 (60) & 1.5 $\pm$ 2.5 & 0.3 $\pm$ 2.9 & 5.6 $\pm$ 7.5 & 4.8 $\pm$ 8.0 & 6.3 $\pm$ 3.9 & 5.3 $\pm$ 5.2 \\
MTF: D4 - D & 10 (90) & 2.1 $\pm$ 1.9 & 0.7 $\pm$ 2.8 & 6.5 $\pm$ 4.8 & 5.8 $\pm$ 5.6 & 6.0 $\pm$ 4.7 & 5.0 $\pm$ 5.8 \\
MTF: D5 - L & 5 (41) & 3.4 $\pm$ 5.2 & -1.5 $\pm$ 6.1 & 9.5 $\pm$ 6.7 & 7.8 $\pm$ 8.6 & 9.9 $\pm$ 5.9 & 7.7 $\pm$ 8.7 \\ \hline
\textit{        MTF subtotal} & \textit{93 (665)} & \textit{2.1 $\pm$ 3.1} & \textit{-0.5 $\pm$ 3.8} & \textit{6.0 $\pm$ 5.6} & \textit{4.4 $\pm$ 6.9} & \textit{7.6 $\pm$ 8.0} & \textit{5.9 $\pm$ 9.4} \\ \hline
\textbf{TOTAL} & \textbf{104 (1040)} & \textbf{3.8 $\pm$ 8.8} & \textbf{-2.5 $\pm$ 9.2} & \textbf{19.9 $\pm$ 30.8} & \textbf{-0.4 $\pm$ 36.7} & \textbf{9.9 $\pm$ 9.7} & \textbf{2.4 $\pm$ 13.6} \\ \hline
\multicolumn{8}{l}{\begin{minipage}[t]{0.8\textwidth}There were no samples for Day 5 Dinner. ``Food volume error'' is equivalent to ``Mean error bias''; ``Error, intake'' is equivalent to ``Volume intake error''; and ``3D \% intake error'' is the same as in~\cite{pfisterer2021segmentation}. RTF: Regular Texture Foods; MTF: Modified Texture Foods; B: Breakfast, L: Lunch, D: Dinner; D\#: Day number. For example, MTF: D1 - L is Modified Texture Foods Dataset: Day 1 - Lunch.\end{minipage}}
\end{tabular}
\bigskip
\bigskip
\caption[Theoretical completion time]{Summary of length of time required to complete food and fluid intake charting for one neighbourhood (unit) comprised of 16 residents (Stage 1) compared to theoretical AFINI-T processing.}
\centering
\label{tab:benchmarked}
\begin{tabular}{lllll}
\hline
\textbf{Type} & \textbf{\begin{tabular}[c]{@{}l@{}}Mode Time \\ (n/N responses)\end{tabular}} & \textbf{Time Range} & \textbf{\begin{tabular}[c]{@{}l@{}}AFINI-T Estimate \\ (1 sec acquisition)\end{tabular}} & \textbf{\begin{tabular}[c]{@{}l@{}}AFINI-T Estimate\\ (10 sec acquisition)\end{tabular}} \\ \hline
Food (per meal) & 10 - 14 mins (3/9) & <10 to 25+ mins & 2 min 34 sec & 9 mins 45 sec \\
Fluid (per meal) & 10-14 mins (4/10) & <10 to 25 mins & N/A & N/A \\
Snack (per snack) & <10 mins (5/9) & <10 to 19 mins & 52 sec & 3 mins 15 sec \\ \hline
\multicolumn{5}{l}{\begin{minipage}[t]{0.8\textwidth}* n is the number of responses with the mode rating out of N, the total number of responses.\end{minipage}}
\end{tabular}
\end{sidewaystable}

\subsubsection{Validating nutrient intake from volume with mass}
\label{resultsvolvsmassvalidation}

\begin{figure*}
	\centering
		\includegraphics[width=.7\linewidth]{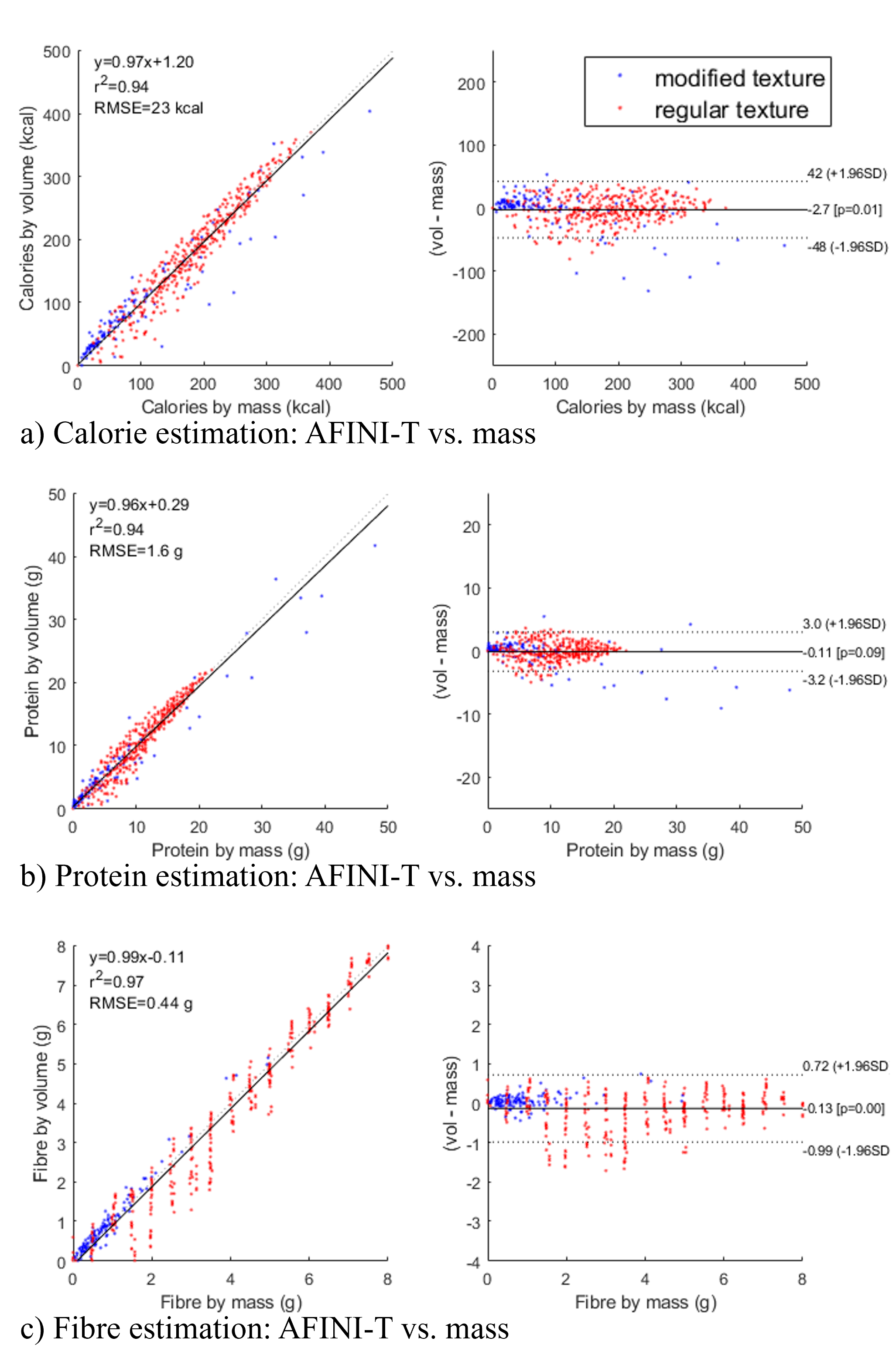}
		\caption[Nutrients of interest correlation and agreement between mass and volume estimates.]{Correlation and agreement between mass and volume estimates for determining nutritional intake at the whole plate level across all imaged samples. Left depicts the goodness of fit with linear regression and coefficient of determination ($r^2$), right depicts the degree of agreement between measures and bias from the Bland-Altman method. Shown here, correlation and agreement between mass and volume estimates of macronutrients: (a) calories, (b) protein, and (c) fibre. Three nutrients of interest are shown here for brevity. For additional nutrient accuracy, see \textit{\textbf{Supplementary Tables} S2, S3, and S4.}}			
		\label{fig:corr_BA_top3}
\end{figure*}

In Figure~\ref{fig:corr_BA_top3}, the MTF plates (blue) tended to be of lesser mass than the RTF (red) largely due to the nature of regular texture foods, representing food choices from the LTC home, were prepared by a supermarket which may not be consistent with LTC serving sizes, whereas MTF were prepared by the LTC home. This translates to a clustering effect of MTF foods at lower values of nutrients with RTF foods towards higher values of nutrients. We also observe a banding effect on fibre for the RTF dataset due to how mass was controlled for matching 25\% portion increments and given the relatively few foods that contained fibre in the RTF dataset. Looking to the spread of nutrient distributions, there is also much higher variance for the MTF dataset for larger amounts of a nutrient (e.g., protein, fat, iron) with tighter variances observed on smaller portion sizes. 

Based on the coefficients of determination shown in Figure~\ref{fig:corr_BA_top3}, nutrient estimates by volume were tightly linearly correlated with nutrient estimates from mass with $r^2$ values ranging from 0.92 for fat to 0.99 for vitamins C, and K. This was true for all nutrients of interest (see \textit{\textbf{Supplementary Tables} S2--S4} for a comprehensive assessment). Based on the Bland-Altman plots, not only were they tightly correlated, there was also good agreement between methods as evidenced by small bias ($\lvert\mu\rvert\leq$  2.7 ) and zero contained within the limits of agreement. Ideally, the bias distributions would be centred around the y-intercept (i.e., a $\mu$ of 0). This was the case with $\mu$ ranging from a minimum of -0.01 for vitamin B6 (mg), zinc (mg), and fat (g), to a maximum of -2.7 for calories (kcal). Taken together, these results suggest nutrient estimation using the AFINI-T system appears to be valid. Estimates were well-aligned with gold-standard weighed food method with the advantage of only a single image acquisition and no need for weighing plates.

\subsubsection{Benchmarking the AFINI-T approach with current practice and requirements}
Let us now compare the end-to-end AFINI-T system against the current workflow. One requirement identified in~\cite{pfisterer2019}, was for the system to run on a portable tablet with inconsistent Wi-Fi. By design, methodology and models were selected to support portability. For example, having selected an approach to support offline training in the EDFN and autoencoder, only the final model residing on the device which does not require Wi-Fi. The autoencoder, which requires only a single training session for global feature extraction, encompasses 84,176 parameters. The per-meal classification layer requires an additional $15n_c$ trainable features, where $n_c$ is the number of food classes for meal $c$ (see Table~\ref{tab:avg_seg_class}). The EDFN food detection network requires 13.7~M parameters, but does not require fine-tuning and can be used globally across meals. 

The second benchmark is with respect to theoretical task completion time. In terms of benchmarking theoretical task completion time, we can compare to results from~\cite{pfisterer2019}. When assuming a very conservative estimate including food handling of 10 seconds per image for acquisition, the time for preprocessing (e.g., plate finding) takes approximately 2.5 seconds per image, with segmentation taking 0.7 seconds per image and classification of 0.05 seconds per image (Dell XPS 15 9570, i7-8750H 2.20GHZ 6-core CPU, NVIDIA GeForce GTX 1050 Ti). As shown in Table~\ref{tab:benchmarked}, even based on these conservative estimates, the theoretical completion time using AFINI-T meets the low end of task completion times (9 minutes 45 seconds versus a mode of 10-14 minutes of completion time for charting one meal). Here, we have assumed separate imaging for each of the appetizer, main and dessert for each resident. If instead we consider acquisition as only acquiring the image (estimated 1 second), this drops to 2 minutes 34 seconds. The true completion time will likely take between these upper and lower bounds, but the key point is AFINI-T is platformed to take less time than the current methodology, and with the added benefit of being objective and capture data at a resident-centric level. Instead of a resident's intake being binned into the 25\% bin across the average foods served that day, AFINI-T captures details to the mL, and tracks personalized items ordered on a resident-by-resident basis.

\section{Discussion}
\label{discussionseg}


While we were able to provide complete total automation for segmentation, classification, and food and nutrient intake estimation for simulated intake plates, the current system relies on the user to specify the current system relies on one fully hand-labelled and hand segmented image for a full reference portion to provide specific semantic segmentation and classification labels for each meal-specific classifier. If this would be deemed as too cumbersome, the next level of automation would be incorporating a semi-automatic method (e.g., graph cut as in~\cite{pouladzadeh2014}) on the backend to aid the user in hand segmenting the reference images. In that case the dietitian or user does not need to hand segment anything, they would simply assign a label to each of the pre-segmented types of foods present with a few lines to indicate where each food type resides (as opposed to hand segmenting an outline for each segment containing every type of food and its associated food-name label). This latter approach is in line with the co-designed user interface and workflow we reported in our previous work where users described acceptability for clicking on a large food region and defining its contents from a drop-down list~\cite{pfisterer2019}. In an LTC setting, this drop-down list could be pre-populated based on the menu items of the day, further simplifying this approach.

Results outlined in Figure~\ref{fig:corr_BA_top3} (and \textit{\textbf{Supplementary Materials} Tables S2--S4 and Figure S1}) demonstrate that the AFINI-T method for estimating food intake is in strong agreement and tightly correlated with true intake. Especially in the case of larger intake portions, the AFINI-T method yielded accuracy of nutrient content with less than 5\% error. For context, comparison to current visual assessment methods with errors in portion size 56\% of the time for immediate estimation and as low as 62\% for delayed recording and stating that current methods error is too high for accurately identifying at-risk residents~\cite{castellanos2002}. Interpretation of the acceptability of the precision and accuracy of the system requires further input from users, ideally using the system in real-world contexts. If warranted, improvements will require a degree of human input or expanded models. This may be in the form showing output classification masks so misclassified segments could be reclassified as appropriate. Alternatively, it could be used to seed food item regions to tightly constrain food regions and then apply region growing to ``intuit'' food segments; this approach is again in line with what was integrated into the collaborative co-design prototype development outlined in~\cite{pfisterer2019}. While not fully automatic, collaborative segmentation through machine learning estimation that is checked and corrected, if necessary, by a human using a simple and intuitive interface would likely be an improvement on current food charting methods, particularly with respect to accuracy, and perhaps time as well. Timed comparison trials would be required to confirm.

\subsection{Comparison to the literature}
It is challenging to assess how AFINI-T compares to the literature because there are no food intake datasets on which to conduct benchmark tests. Additional considerations affecting the ability to compare include the number of included classes, inconsistencies with ``accuracy'' reporting (e.g., top-1 vs top-4 accuracy) and the complexity of the classification problem (e.g., whole raw foods vs prepared meals vs modified texture versions of those prepared foods). Some accuracy for classification methods based on handcrafted features has been reported in the literature: Zhang \textit{et al.} report 88.2\% accuracy for whole foods (entire pineapple) consisting of 18 classes~\cite{zhang2012}, Bolle \textit{et al.} achieved 95\% using top-4 accuracy for vegetable identification in a supermarket~\cite{bolle1996veggievision}, Rocha \textit{et al.} achieved 99\% using top-2 accuracy for some fruits and vegetables by fusing three types of features (including Unser's features)~\cite{rocha2010automatic}, Arivazhagan \textit{et al.} achieved 85\% accuracy on 15 types of produce using a minimum distance classifier~\cite{arivazhagan2010fruit}, and Chowdury \textit{et al.} attained 96.55\% accuracy on 10 vegetables using colour and texture features and a neural network classifier~\cite{chowdhury2013vegetables}. Regarding a trend for learned features, a deep learning approach has had a comparatively slow adoption in the field of food imaging. Accuracy around segmentation and classification tends to be either not mentioned~\cite{chen2012automatic, miyazaki2011, ege2017, fang2015, chokr2017} or was explicitly stated as being beyond the scope of the present version of their system~\cite{miyazaki2011}. In recent years, systems have begun to leverage learned features~\cite{meyers2015,pouladzadeh2016food,ege2017}. Feature extraction and classification are combined, making sources of error difficult to disentangle. This is further confounded when segmentation and classification accuracies are combined instead of considering them as two sub-processes. Classification accuracies using deep learning reported range from  100\% (11 classes)~\cite{pouladzadeh2016food} to 82.5\% (15 classes)~\cite{ege2017}. Alternative methods employed for classification were AdaBoost~\cite{miyazaki2011}, K Nearest Neighbours~\cite{he2013food}, and support vector machines~\cite{chokr2017,chen2012automatic} with reported classification accuracies of 68.3\% (50 classes)~\cite{chen2012automatic} to 99.1\% (6 classes)~\cite{chokr2017}. While direct comparison between the AFINI-T system and other automated methods for assessing LTC intake data is not possible since the AFINI-T system is the first to measure food intake and considers modified texture foods, these results suggest that AFINI-T's DNN approach is among the highest performing approaches with a top-1 accuracy of 88.9\%. Furthermore, the type of data represented in the MTF and RTF datasets for LTC contain more complex food scenarios as they are prepared foods (RTF: 93.9\% accuracy; MTF: 73.7\%), and the accuracy we report is top-1, which means the AFINI-T approach may outperform the others.

At the inference level, few papers report percent error at the nutrient level and tend to focus on calorie estimation. The few studies that have reported calorie estimation error of: 0.09\% (mean absolute error) on 6 categories using random forests and support vector machines~\cite{chokr2017}, and 0.25\%  (mean standard error) on 11 categories of entire foods (e.g., green pepper) using a CNN~\cite{pouladzadeh2016food}. Others have reported 80\% of calorie estimates falling within 40\% error (35\% within 20\% error) on 15 classes using a multi-task CNN with a maximum correlation coefficient of 0.81 (r$^2$ 0.64 equivalent), and top-1 accuracy of 82.48\%~\cite{ege2017}. Previous research~\cite{ege2017} also reports a comparison to~\cite{miyazaki2011} with 79\% of calorie estimates falling within 40\% error (35\% within 20\% error) using hand-crafted features with a correlation coefficient of 0.32 (r$^2$ 0.10 equivalent)~\cite{miyazaki2011}. The AFINI-T system demonstrated an error of 2.4\% across 13 nutrients on the 56 categories (102 classes) of food with an minimum r$^2$ value of 0.92 (0.94 for calories). The average top-1 accuracy was 88.9\%, ranging between 95.1\% on 3 classes (RTF: Dinner) to 70.4\% and 89.0\% on 15 class meals (MTF: Day 3 Dinner, MTF: Day 1 - Lunch). Based on these comparisons, this work performs among the best reported in the literature despite having more complex meal scenarios and across 13 nutrients. While there has been relatively little work done in this area, these results represent a novel contribution both from the technical implementation as well as real-world implementation perspectives.

While each of the existing methods identified above have unique strengths and limitations, they do achieve similar accuracies to average reported human error with respect to portion size or calorie estimation. Benefits of the AFINI-T system include that it can measure a specific resident's intake (as opposed to the proportion consumed across the average of all foods offered), with performance at least matching others approaches. Compared to the current visual assessment methods, it is easy to use, fast to acquire and process, removes subjectivity and provides repeatable estimates, and it can be tracked to the nutrient level to provide a comprehensive profile of each resident-specific intake in a quantitative way. This translates into higher quality data that could be used to inform resident preferences and streamline referrals to registered dietitians along with a data-driven approach for monitoring and evaluating nutritional interventions.

\subsection{Disentangling sources of error for nutrient estimation}
While evaluation of accuracy in real-world contexts is prudent, with the current AFINI-T approach we show segmentation of only one reference image is required and that even when some pixels are misclassified, there is reasonable nutrient intake accuracy robustness. These misclassifications tended to occur near the edges of a food segment regardless of dataset which may be due to a less uniform representation near the edges either due to higher crumbliness (e.g., meatloaf crumbs), or due to the convolutional kernel extending into the "empty space" (i.e., the plate) making it easier to classify a pixel as food when there are food pixels surrounding it. Should further improvements be desired, human-in-the-loop for rectifying segmented region errors may be necessary, but as we saw in Section~\ref{resultsvolvsmassvalidation}, nutrient intake estimation relative to ground truth weighed-food records indicate these errors in classification of some regions do not appear translate to large intake errors and this fully automated classification strategy may be deemed acceptable given the time-savings. Additional consideration for technology translation is warranted. 

Another consideration is that not all foods have nutrient values for every nutrient of interest and often rely on complex imputations for estimates~\cite{ispirova2020evaluating}. For example, nutrients like vitamin D are often incomplete~\cite{holden2008assessing}. Additional discussions with end-users and nutrition experts are warranted to evaluate the utility and appropriateness of reporting these values, what margin of error is deemed acceptable for supporting trust in the system, and other considerations given the quality of data included in the underlying nutritional databases.

As observed in Figure~\ref{fig:corr_BA_top3}, with smaller intake amounts and therefore smaller relative portion differences, the error was larger. To improve on this in the future, we must consider from where this error arose. One contributor may be depth map variance. Results indicate that nutritional intake estimates, had greater variation at the lower levels of intake (larger spread at lower intake levels). This may correspond to the amount of variation in estimation at smaller levels of intake/larger food left on the plate. We speculate this is because of a compounding of small discrepancies in depth maps which gets propagated to volume and then to nutritional intake. Future work will address this by incorporating depth-map variance as a feature to describe the food item. For example, for a green salad, we would expect a higher variance in depth map because it is a non-dense food item and in contrast, meatloaf or slab cake would have very low depth-map variance across the food item as these items are more block-like. Exploring automated 3D segmentation may also be intriguing where depth information could be stacked onto color channels and thus incorporated into salient feature extraction. Similar approaches have shown promise in recent advances in agriculture~\cite{xia2015situ,lin2020color,gai2020automated}, construction~\cite{beckman2019deep}, and robotics and automation~\cite{danielczuk2019segmenting}.

The datasets collected as part of this work (as described in~\cite{pfisterer2021segmentation}) were the first semantic food segmentation datasets representative of LTC foods, with pixel-wise segmentation and depth map information, and at multiple simulated intake levels. These provide a foundation upon which to grow. Computationally simulated plates for modified texture foods on a plate by combining modified texture single food items from the modified texture food dataset into computationally simulated multi-item plates. This may provide additional insights into performance and accuracy more closely resembling real-world data. Furthermore, additional expansion to test the system with a wider variety of foods with fewer constraints may enhance generalizability, which in turn would support greater accuracy and encourage uptake of this technology. For example, the level of plate mixing in a real-world setting will have greater variability than what was tested in this research, thus it must be explored and evaluated as part of future work. 

Using the current system, low-density foods (e.g., salad), rigid foods (e.g., toast), and highly absorptive foods (e.g., blueberry) continue to pose challenges for RGB-D imaging. Empirically, most of the misclassified pixels were around the edges of a food segment. In this region, the foods had a higher variance in terms of their representation. We hypothesize that increasing the size of the training dataset and including representations of ``left overs'' instead of only full portions could help learn a more generalizable classifier.

While we collected ground truth weighed food records, we did not account for ground truth volume. As a result, we were working under the assumption that AFINI-T's volume assessment was accurate. Volume validation against gold-standard ground truth (e.g., water displacement) is be needed to corroborate the accuracy (although in actuality, some evidence suggests there is less than 3\% volume error of the RealSense~\cite{liao2016}). This is an important consideration for more thoroughly quantifying error at each stage. Given the state of the literature in how error is typically reported (if it is at all), this paper provides a step towards more transparent technology for supporting trust in the system.

From a translational standpoint, AFINI-T is platformed to provide actionable data-driven insights that could help inform menu planning by dietitians and director of food services. For example, it could be used to develop recipes that are more nutrient dense that complement the nutrients in recent past meals. Creating nutrient dense meals while minimizing cost is a priority in LTC as there is a fixed allocation of food cost per resident. The raw food allocation in Ontario was \$9.54 per resident per day in 2020~\cite{OLTCA2020}. Until recently, there was a disconnect between the perceived requirement to serve full portion to meet nutritional requirements (i.e., the portion size which was costed to provide adequate nutrition); however, because of limited budget, the foods that were served were relatively inexpensive and the quantity required was unsuitable. This resulted in a high degree of food waste~\cite{duizer2020menuplanning,grieger2007nutrient} while increasing risk for malnutrition due to less consumed nutrients than the planned nutrient consumption~\cite{keller2018prevalence}. AFINI-T could also be used as a tool for developing more nutrient-dense recipes in which certain ingredients could be replaced with others. For example, replacing half of the ground beef in a chili recipe with lentils to decrease saturated fat and cholesterol while increasing fibre. Data on which foods are consumed can inform how to design recipes to be smarter, more expensive but also more nutrient dense, with the expectation of less waste and more of portion consumed, especially when paired with software such as Food Processor for designing recipes. While these types of strategies were not part of this work, they are direct motivation for it and have great potential to impact and disrupt the way we assess nutrition management and beyond when they are explored in future work.

Future directions also include adding an additional stage for automatic food-type classification as specific foods rather than arbitrary classes with associated nutritional values (i.e., mashed potatoes are classified as mashed potatoes after the initial segmentation step). A human-in-the-loop version where there is the opportunity to correct all misclassified regions (i.e., the best-case scenario) could further improve results, albeit at the expense of manual hands-on time and effort, which needs to be minimized. In addition, improving upon the algorithms to handle more complex food types (e.g., salads and/or soups in which the food is comprised of multiple components) as well as more complex plates of food to address food mixing as seen with mashed potatoes will improve AFINI-T's ability to assess plates ``in the wild".

\subsection{Conclusion}
\label{conclusion}

AFINI-T is a deep-learning powered computational nutrient sensing system that provides an automated objective and efficient alternative for food intake tracking providing food intake estimates. Novel contributions of this approach include: a novel system with decoupled segmentation, classification, and nutrient estimation for monitoring error propagation, and a convolutional autoencoder network for classifying regular texture and modified texture foods with a top-1 accuracy of 88.9\% with a mean intake error of 0.4 mL and nutritional intake accuracy with strong agreement with gold standard weighed food method and good agreement between methods ($r^2$ from 0.92 to 0.99; $\sigma$ range from \textminus2.7 to \textminus0.01; zero within the limits of agreement) across 13 nutrients of interest to LTC. Translation of AFINI-T may provide a novel means for more accurate and objective tracking LTC resident food intake providing new resident-specific insights for supporting and preventing malnutrition. AFINI-T's data-driven insights may streamline and prioritize dietitian referrals for supporting nutritional intervention efficacy. This may enhance sensitivity of catching at-risk residents and enable more holistic monitoring for malnutrition reduction.

\section{Methods}

\subsection{End user data to shape technological requirements: A case study}

Insights motivating the technical approach described in this paper were gathered through interviews and workshop discussion with Schlegel Village team members collected during, but not included in our previous user study paper~\cite{pfisterer2019}. Two interviews (a registered dietitian (RD) nutrition research expert, and an RD working in LTC), and discussion with experts during a workshop were conducted. The workshop included 21 participants representing 12 LTC and retirement homes who were recruited through self-enrollment including an Administrative Assistant, Chef, Dining Lead (similar to a dining room manager), Director of Recreation, Dietary Aides, Neighbourhood Coordinator, Recreation Assistant, Restorative Care, Senior Nurse Consultant, Directors and Assistant Directors of Food Services, Registered Nurse, and Personal Support Workers (PSW)~\cite{pfisterer2019}. Participants identified potential barriers to uptake including time, and whether the level of detail is desired or seen as valuable.

\subsubsection{Summarizing user needs to inform design requirements}
Qualitative results from interviews and workshops with end-users illuminated the following user needs which, guided by grounded theory~\cite{chun2019grounded}, were translated by into design requirements (described in Section~\ref{Results - Case Study}) for application within the LTC context.
\begin{itemize}
        \item \textbf{\textit{Nutrients of concern}}: 13 nutrients of concern were identified based on clinical endpoints and expert user feedback. \textit{Macronutrients}: calories, carbohydrates, fats, fibre, and protein. \textit{Micronutrients}: calcium, iron, sodium, vitamin B6, vitamin C, vitamin D, vitamin K, and zinc.
        \item \textit{\textbf{Dietitian's role}}: Guided by current workflow, the system must record accurate information which can then be leveraged by the dietitian to prioritize referrals and more easily manage nutritional interventions as well as the complex considerations for what level of information to be provided to which level of care (e.g., dietary aide versus dietitian). The dietitian must be the gatekeeper.
        \item \textit{\textbf{Real-time use}}: Guided by current workflow, an on-board system without constant internet access dependency informs how classification models are trained and updated. 
        \item \textit{\textbf{Simplifying classification}}: Given the identified time-constrained nature of LTC, to address concerns for a streamlined system, we leverage mealtime and prescribed diet texture as \textit{a priori} information to constrain the classification problem to minimise potential barriers to uptake a level of automated segmentation and classification~\cite{pfisterer2021segmentation}. However, the challenge with even the most successful of generic classification methods is they fail to reach 100\% accuracy which may limit their utility in practice especially if classification does not include assessment of food (or nutrient) intake accuracy. Within nutrition tracking in long-term care, this can have very real consequences. If mayonnaise gets misclassified as low-fat Greek yogurt for example, the nutritional composition recorded may be drastically affected, causing negative implications for both understanding nutrient intake as well as staff's trust in the system. When considering the necessity for accurate nutrition tracking, we therefore need a solution that performs reasonably well and acknowledge some degree of human assessor intervention may be required if error-margins are inappropriately high. By (a) incorporating \textit{a priori} information about when to expect which foods (e.g., LTC menu), and (b) leveraging prescribed diet for further refinement, we can limit the extent to which human intervention is required. For example, if we know Mrs. Brown takes her chicken minced, then today's minced lunch classifier should be used for Mrs. Brown's data. 
\end{itemize}

\subsection{Experimental procedure: AFINI-T's technical approach}
\label{ssec:procedureseg}
\subsubsection{Data collection}
As described in~\cite{pfisterer2021segmentation}, data were collected in an industrial research kitchen at the Schlegel-University of Waterloo Research Institute for Aging's Centre of Excellence for Innovation in Aging. This kitchen was modeled after industrial research kitchens found in LTC homes. An optical imaging cage was constructed that enabled top-down image capture as described in~\cite{pfisterer2021segmentation}. The camera was connected to a computer for data acquisition and plates were weighed at a nearby weigh station.

\subsubsection{Regular and modified texture food datasets}
We used our regular texture foods dataset and our modified texture foods dataset. Table~\ref{tab:dataset_overview} provides an overview of dataset characteristics and a summary of all food items imaged can be found in \textit{\textbf{Supplementary Table} S1}. For each food item, one full serving was defined by the nutritional label portion size (regular texture dataset) or the recipe-defined portion size received from the kitchen, and was weighed to the nearest 1 gram using an Ohaus Valor Scale.  

\begin{table}[]
\centering
\caption[Dataset characteristics.]{Overview of dataset characteristics. The regular texture food dataset (RTF) was comprised of 3 ``meal'' plates each consisting of 3 foods imaged at every permutation of 25\% simulated intake. The modified texture food dataset (MTF) set consisted of 134 food samples representing 47 foods each consisting of a set of at least one pur\' ee and one minced texture food. Each sample was imaged 5 times by progressively removing food with the exception of 6 samples consisting of 4 each with one lost image.}
\label{tab:dataset_overview}
\begin{tabular}{llll}
\hline
Dataset overview & \textbf{RTF} & \textbf{MTF} & \textbf{Total} \\ \hline
\# images             & 375 & 664 & 1039 \\
\# samples            & 3   & 134 & 137  \\
\# classes            & 9   & 93  & 102  \\
\# foods represented  & 9   & 47  & 56   \\
\# foods with recipes & 9   & 27  & 36   \\ \hline
\end{tabular}
\end{table}

For the \textit{\textbf{regular texture foods dataset}}, where a serving size was referenced using volume, that volume of food (e.g., corn) was weighed and the mass was used thereafter. Nutritional information for each food item was supplied by the manufacturer except for meatloaf and mashed potatoes, for which the nutritional information supplied by the manufacturer was combined for both food items and not on the individual food item level, so nutritional information was approximated using USDA's food database. Since manufacturers supply nutritional information for minerals as percent daily value (assuming a 2000 calorie diet), for the RTF dataset, minerals were reported similarly. For more details on the version, see \textit{\textbf{Supplementary Table} S5}. Mass in grams was used to define all serving sizes.

For the \textit{\textbf{modified texture foods dataset}}, we expanded our original MTF dataset~\cite{pfisterer2021segmentation} with additional examples (that did not have recipes) for further segmentation and volume estimation analysis. The nutritional analysis was conducted on the subset of the 314 images. To approximate a true food consumption more closely, the modified texture food dataset was less stringently controlled in terms of portion size. The portion size was defined as the amount recorded on the LTC kitchen's recipes (defined in millilitres). Since nutritional information were provided according to a volumetric serving size, we needed to convert from mass to volume for using weighed food for nutrient intake validation to define the expected mass of a full portion in millilitres. To accomplish this, we calculated the food's density to convert by using the full plate’s “true volume” (in mL) with its mass (in grams). This enabled us to scale nutritional information using the same pipeline as the regular texture foods dataset for validating these findings using mass; it was not required for the system to operate.

\subsubsection{Training dataset}
\begin{figure}
	\centering
	\includegraphics [width=.8\textwidth]{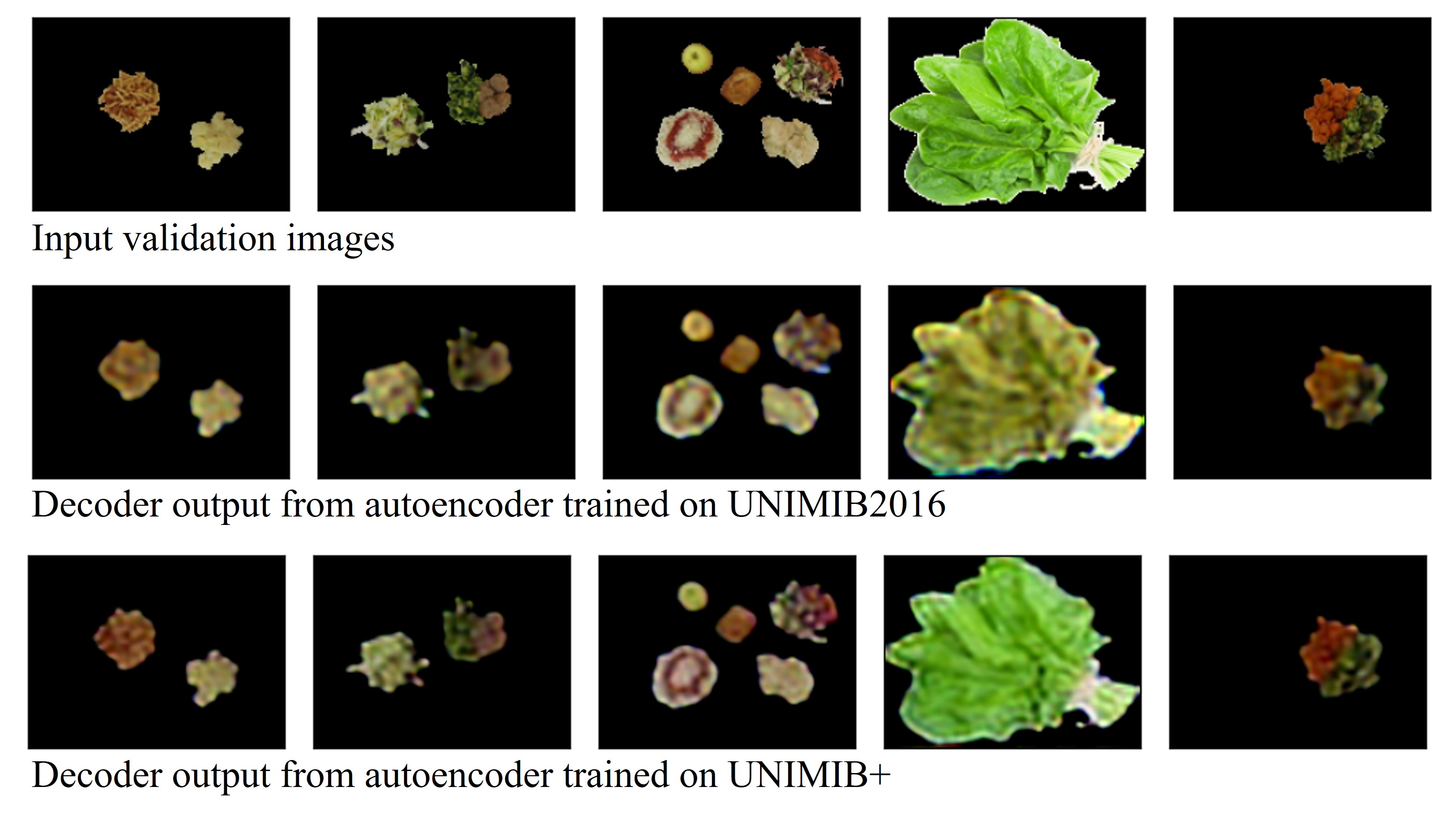}
    \caption[Effect of underrepresentation of green foods on decoder output.]{Effect of underrepresentation of green foods on decoder output. The decoder output from the autoencoder trained on the UNIMIB+ dataset in the bottom appears less murky, more vibrant, and with truer perceived greens than the UNIMIB2016 counterpart in the middle.}
\label{fig:unimib_vs_unimib+}
\end{figure}

We used the UNIMIB2016 dataset to train the convolutional autoencoder (described in detail in Section~\ref{sec:comp_methods} with system diagram in Figure~\ref{fig:vol2nuts_SystemDiagram}). However, we discovered that UNIMIB2016 had an underrepresentation of green foods compared to what is served in LTC. This affected the autoencoder's ability to differentiate between all colours and textures. To address this difference in the canteen/cafeteria images from the original UNIMIB2016, we augmented the training dataset by adding in 91 examples of lettuce, 91 examples of peas, and 89 examples of spinach from the FoodX-251 food dataset~\cite{kaur2019foodx}. We refer to this as the UNIMIB+ dataset. Figure~\ref{fig:unimib_vs_unimib+} shows the effect of this underrepresentation of green by its inability to reconstruct a vibrant hue across the autoencoder's decoder output trained solely on the UNIMIB2016 dataset for validation examples. The autoencoder was able to converge to a lower validation loss on the UNIMIB+ dataset. Empirically, this resulted in greens appearing greener, reds appearing redder, yellows and whites appearing less murky as well, as shown in the bottom UNIMIB+ examples compared to the UNIMIB2016 in Figure~\ref{fig:unimib_vs_unimib+}. This suggests the addition of the green samples enabled the autoencoder to learn good food representations and encode features more deeply and were aligned more closely with how a human would perceive the foods; a crucial point for the LTC application.

\subsubsection{Computational methods}
\label{sec:comp_methods}

The following describes how the segmentation strategy was refined compared to our initial work in~\cite{pfisterer2021segmentation}, the general food/no-food classification approach, followed by system automation through the use of a convolutional autoencoder.

\bigskip
\noindent {\textit{\textbf{Refined segmentation strategy}}}

\noindent Modifications to the training process were made to enhance network performance. In general, model overfitting happens when a model becomes too specific to the training set, at the expense of generalizability to other data. We introduced early stopping criteria to halt training early to avoid this overfitting problem, yielding a network that was trained over fewer epochs than one that is overtrained. This yielded an image mask that classified each pixel as food or no-food with calibrated depth~\cite{pfisterer2021segmentation}. Comparisons of volumes were made to infer volume of food consumption from the relative difference to the full serving reference portion~\cite{pfisterer2021segmentation}. Volume consumed was mapped onto nutritional information for intake approximation. These nutrient level intake estimates were then validated against the ground truth nutritional information obtained through the weighed-food method.

\begin{sidewaysfigure*}
	\centering
	\includegraphics [width=1\textwidth]{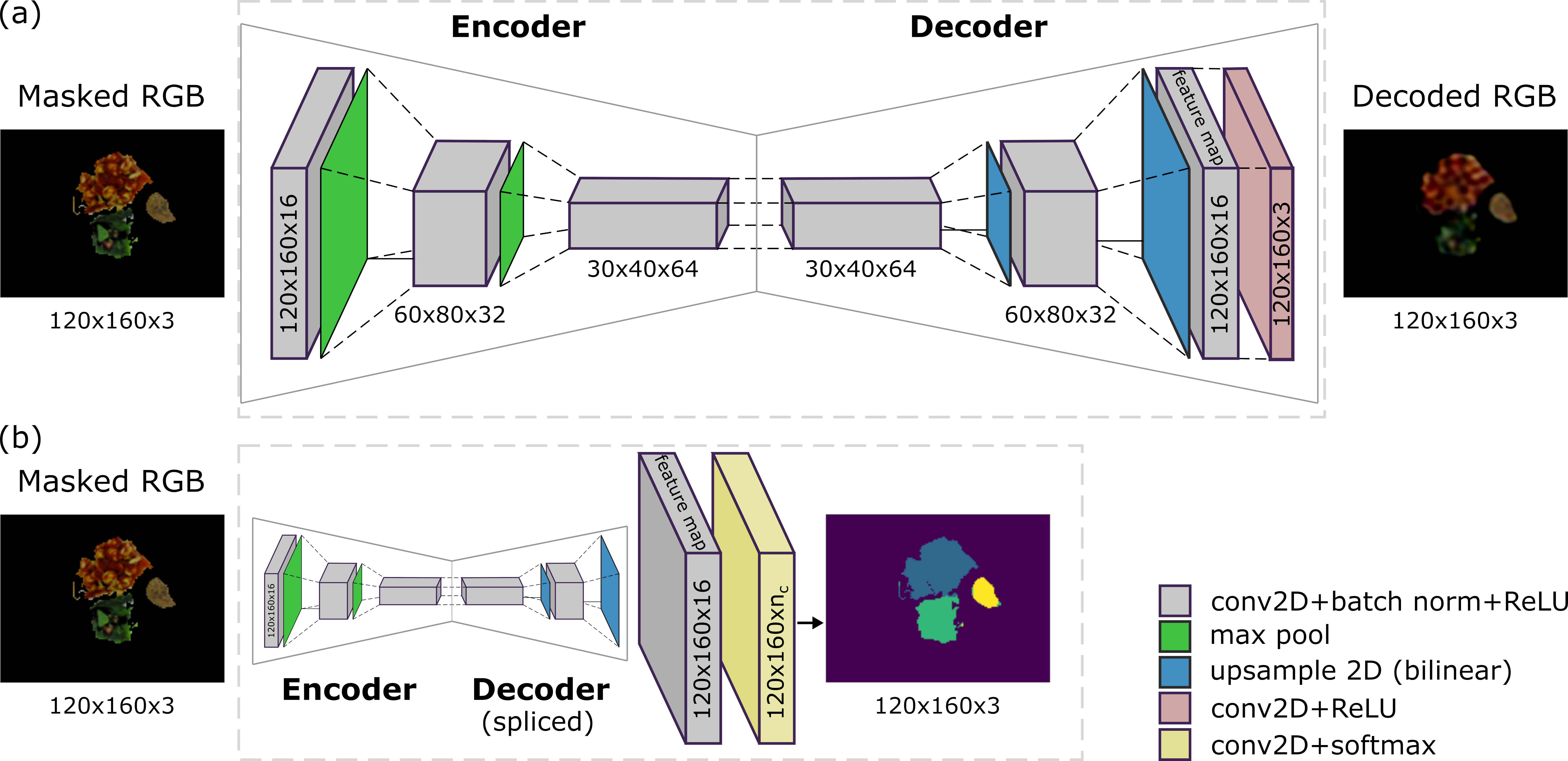}
    \caption[Convolutional autoencoder network for learned feature representation and in the context of classification.]{Convolutional autoencoder network for learned feature representation and in the context of classification. (a) The architecture for learning feature representation: an input image is given and the output is a reconstruction of that image. Training minimized the error between input and output images; we used a MSE loss with Adam optimizer, a learning rate of 0.0001, and a batch size of 32. The early stop criteria used were a change of loss of $<$ 0.0001, and a patience of 5 epochs. (b) The autoencoder was spliced, weights frozen and only a classification layer for $n_c$ classes was trained for classification, where $n_c$ is the number of food items for meal $c$. We used categorical cross-entropy (ignoring background pixels) loss, with Adam optimizer, and a learning rate of 0.1. The early stop criteria used were a change of loss $<$ 1x$10^{-5}$, and a patience of 5 epochs. We used a 70\%:30\% train to validation split of augmented data. The data were augmented by generating 300 images from the full set of plates and applying random flips, rotations, and increased or decreased contrast.}
\label{fig:vol2nuts_SystemDiagram}
\end{sidewaysfigure*}

\bigskip
\noindent {\textit{\textbf{General classification approach}}}

\noindent Here the UNIMIB+ data was used and spliced off the autoencoder at the feature map as a latent feature extractor for classification (see Figure~\ref{fig:vol2nuts_SystemDiagram} for a system diagram and network architecture). We took a similar approach to our previous research on to classification for predicting relative nutritional density of a dilution series of commercially prepared pur\' ees~\cite{pfisterer2018}. There we trained a global autoencoder across flavours and fine-tuned the model for flavour-specific modules and compared against hand-crafted features (64 colour; 7 texture) and the learned features. Given the similarity between this dilution series and modified texture foods which comprise 64\% (664/1039) of our testing dataset and 47\% of the LTC population receives modified texture foods~\cite{vucea2019prevalence}, taking a similar approach seemed appropriate. 

\bigskip
\noindent {\textit{\textbf{Automation with a convolutional autoencoder}}}

\noindent We report nutrient intake accuracy using the automated system (i.e., the automated classification case). By incorporating an initial pass of automation, the goal is to reduce the degree of intervention required by the user. For this automated approach, we developed a semantic segmentation network with a convolutional autoencoder feature extractor for classification of foods which was roughly inspired by a highly successful convolutional neural network (CNN), the VGG network~\cite{simonyan2014}. For a given meal or time of day, we fed the ``masked'' output from the EDFN-D (food/no-food detector as described in~\cite{pfisterer2021segmentation}) into the convolutional autoencoder. CNNs encode spatial information and given how food has differing degrees of cohesion, we felt the context of spatial information would be an asset. We chose a convolutional autoencoder approach because while we have fully labelled data for these datasets, in practice, this may be infeasible to collect; it will be far more likely that we have a large proportion of unlabelled data. Additionally, we sought to extract latent features via a method which only requires one round of training offline, to be ported over to the working system and then a small classification layer could be appended and trained for each meal using \textit{a priori} information about the meal items offered. Autoencoders can extract meaningful, generalizable features because their framework is built for data compression, which tries to optimize the encoded representation to reconstruct the original input in an efficient manner. It is comprised of an encoder which learns descriptive features in a compressed way, and a symmetric decoder, to reconstruct the estimated signal. Loss was computed as pixelwise mean squared error between the input and reconstructed output; as a result, they do not require labelled training data. They also provide an advantage in interpretability because you can visualise the reconstruction and observe its behaviour to investigate network behaviour. As we saw in Figure~\ref{fig:unimib_vs_unimib+}, this characteristic can be incredibly valuable for understanding inherent bias of training datasets. The ``convolution'' refers to learning spatial patterns where certain features exist over an image.

We trained an autoencoder to be a feature extractor using the UNIMIB+ dataset, consisting of 1277 images. Data were split 70\% training and 30\% validation. Training was performed using the Adam optimizer with batch size of 32, mean squared error loss and early stopping (<0.0001 validation loss change) with 5-epoch patience. Only food pixels were used in the loss calculation, using the ground truth masks. The convolutional autoencoder network was spliced before the final 1x1 convolution block to produce original resolution 16-channel latent feature vectors. The weights of this network were frozen and used as a feature extractor for the classification training.

We trained a different classifier for each meal (described in more detail below). In an LTC environment where the menu offerings for each meal are pre-determined, the classification problem can be constrained to each meal with \textit{a priori} known possible options. Given that there are many food options and as new meals are planned, we needed to adopt a flexible and modular approach given the realities of multiple food choices each day, which also enables us to use only one labelled example per item. Using the AFINI-T method, only one full reference portion is required for comparing leftover plates against to classify foods and infer intake. For nutritional intake estimation, we leveraged nutritional info per portion based on the Schlegel Villages menu planning software (or supplied by the manufacturer in the regular texture dataset) to link up proportional nutrient intake with the portion of food under the assumption that recipes were followed exactly. 

Denoting the number of menu items for meal $m$ as $n_c$, the classification network for meal $c$ was built by appending $n_c$ 1x1 convolution kernels onto the feature extractor network. By freezing the feature network, per-meal training was fast, requiring only optimizing the final layer’s weights. The training data was constructed by augmenting the full set of plates by applying random flips, rotations, and increased or decreased contrast. We used one reference image (the full-portion image) to learn what each class looked like and then mapped subsequent instances onto these pre-labelled classes by grouping all the ``full plates'' of food for a given meal into the training set. The data were split into 70\% training and 30\% validation. Training was performed using the Adam optimizer with batch size of 32, categorical cross-entropy loss and early stopping (<1x$10^{-5}$ validation loss change) with 5-epoch patience. Only food pixels were used in the loss calculation, using the ground truth masks.

We applied ground truth labels to the full portion plate so we could link up the proper proportional intake at the nutrient level and assess accuracy of the intake estimates including how this image-based system compares to the gold-standard weighed food approach. These classes could be left general for bulk intake estimates that are class-agnostic and would require no user input, in which case it would assume equal consumption across a plate, similar (but less subjective) to what is currently being tracked in LTC. For example, lunch contained 6 items (classes) and some food pixels are classified as “class 1”. While the network would not know which food comprises class 1, it knows they ate 79\% of the plate that was served to them. Here, we instead chose to incorporate an additional level of detail (i.e., class 1 is ``meatloaf'') for more accurate nutrient intake assessment to create a system which allows for flexibility in the level of detail it provides. For example, there may be scenarios when bulk intake estimates are preferred for their increased efficiency, while at others, dietitians would prefer a more fine-grained approach for a subset of at-risk referred residents.  In this case the automated approach (with possible human-in-the-loop refinement) provides an automated (or semi-automated) alternative to support food-item specific intake without the assumption that foods are equally consumed.

\subsubsection{Nutrient intake association}
\label{sssec:nutrition}
This step was comprised of three general stages: (1) determine the relative consumption of each food item compared to a full reference portion using food volume estimation from the depth maps, (2) compare relative consumption to nutritional information to infer nutritional intake for each item, and (3) sum the inferred nutritional intake for each item across a plate for an estimation of total nutrition consumed during a meal (for the modified texture foods, this was across the plate of one food item). More specifically,

\begin{enumerate}
\item \textbf{Use relative changes in volume to estimate food intake:	}
The 0\% eaten portion for each food item was considered as the full portion. Each subsequent portion (i.e., 25\%, 50\%, 75\%, 100\% consumed, or portions 2, 3, 4, 5 for the MTF dataset) for each food item was compared against the initial full portion to yield a relative volume change representing the intake of that specific food item. The relative change in volume was compared to the ground-truth relative change in mass for each food item.

\item \textbf{Use food intake to estimate nutrient intake for each food item:}
The nutritional information for one serving was used as the reference full portion for each of the priority nutrients: calories, carbohydrates, fats, fibre, protein, calcium, iron, sodium, vitamin B6, vitamin C, vitamin D, vitamin K, and zinc. Given the proportion of consumed food for each food item, the nutritional information was scaled accordingly to estimate intake for each of the priority nutrients at the food item level.


\item\textbf{Use food intake to estimate overall nutrient intake:}
Finally, the total nutrient intake consumed during the meal was estimated by summing the nutrient intake across all food items for each plate.	
\end{enumerate}


\subsection{Statistical Analyses}
\label{ssec:stats}
\subsubsection{System accuracy}

Segmentation accuracy was assessed using intersection over union (IOU). Classification accuracy was described using top-1 accuracy and is summarized with per-meal classifiers. Bulk intake accuracy (i.e., class-agnostic, overall food volume intake) was assessed using mean absolute error (mL) and 3D\% intake error also described in~\cite{pfisterer2021segmentation} where intake error was calculated for volume (3D) data relative to the full portion. All values are reported as mean~$\pm$~SD. Nutrient intake accuracy was assessed using the fully automated classification approach (i.e., without updating misclassified regions) to evaluate nutrition intake accuracy and is reported as mean~$\pm$~SD as well as \% error.


\subsubsection{Validating nutrient intake estimation against weighed-food records}

All data were analyzed using MATLAB 2020b (MathWorks, Natick, MA). Linear regression was used to determine the goodness of fit through the degree of correlation with $r^2$ to summarize the extent to which nutritional intake information from weighed-food mass is related to estimated nutritional information from food volume. Bland-Altman analysis was used to describe the level of agreement between nutritional intake info from weighed-food mass compared to intake volume with mean agreement ($\sigma$) and bias ($\mu$) between methods~\cite{giavarina2015}).

Several nutrients of concern in the regular texture foods dataset were reported in \% daily value (i.e., calcium, iron, vitamin B6, vitamin C, and zinc). We converted these to absolute values to match the modified texture foods dataset using the 2005 Health Canada reference values for elements and vitamins. Where there was a difference across age, we used the >70 years old reference; where there was a difference in requirement by sex, we took the average value.

\section*{Data Availability}
Data are available by contacting the corresponding author on reasonable request.



\bibliography{bib}

\begin{thebibliography}{100}
\urlstyle{rm}
\expandafter\ifx\csname url\endcsname\relax
  \def\url#1{\texttt{#1}}\fi
\expandafter\ifx\csname urlprefix\endcsname\relax\def\urlprefix{URL }\fi
\expandafter\ifx\csname doiprefix\endcsname\relax\def\doiprefix{DOI: }\fi
\providecommand{\bibinfo}[2]{#2}
\providecommand{\eprint}[2][]{\url{#2}}

\bibitem{pirlich2001}
\bibinfo{author}{Pirlich, M.} \& \bibinfo{author}{Lochs, H.}
\newblock \bibinfo{journal}{\bibinfo{title}{Nutrition in the elderly}}.
\newblock {\emph{\JournalTitle{Best Practice \& Research Clinical
  Gastroenterology}}} \textbf{\bibinfo{volume}{15}}, \bibinfo{pages}{869--884}
  (\bibinfo{year}{2001}).

\bibitem{keller2004}
\bibinfo{author}{Keller, H.~H.}, \bibinfo{author}{{\O}stbye, T.} \&
  \bibinfo{author}{Goy, R.}
\newblock \bibinfo{journal}{\bibinfo{title}{Nutritional risk predicts quality
  of life in elderly community-living {C}anadians}}.
\newblock {\emph{\JournalTitle{The Journals of Gerontology: Series A}}}
  \textbf{\bibinfo{volume}{59}}, \bibinfo{pages}{M68--M74}
  (\bibinfo{year}{2004}).

\bibitem{lanctin2021prevalence}
\bibinfo{author}{Lanctin, D.~P.} \emph{et~al.}
\newblock \bibinfo{journal}{\bibinfo{title}{Prevalence and economic burden of
  malnutrition diagnosis among patients presenting to united states emergency
  departments}}.
\newblock {\emph{\JournalTitle{Academic Emergency Medicine}}}
  \textbf{\bibinfo{volume}{28}}, \bibinfo{pages}{325--335}
  (\bibinfo{year}{2021}).

\bibitem{keller2017prevalence}
\bibinfo{author}{Keller, H.~H.} \emph{et~al.}
\newblock \bibinfo{journal}{\bibinfo{title}{Prevalence and determinants of poor
  food intake of residents living in long-term care}}.
\newblock {\emph{\JournalTitle{Journal of the American Medical Directors
  Association}}} \textbf{\bibinfo{volume}{18}}, \bibinfo{pages}{941--947}
  (\bibinfo{year}{2017}).

\bibitem{keller2019prevalence}
\bibinfo{author}{Keller, H.} \emph{et~al.}
\newblock \bibinfo{journal}{\bibinfo{title}{Prevalence of malnutrition or risk
  in residents in long term care: comparison of four tools}}.
\newblock {\emph{\JournalTitle{Journal of Nutrition in Gerontology and
  Geriatrics}}} \textbf{\bibinfo{volume}{38}}, \bibinfo{pages}{329--344}
  (\bibinfo{year}{2019}).

\bibitem{bell2013prevalence}
\bibinfo{author}{Bell, C.~L.}, \bibinfo{author}{Tamura, B.~K.},
  \bibinfo{author}{Masaki, K.~H.} \& \bibinfo{author}{Amella, E.~J.}
\newblock \bibinfo{journal}{\bibinfo{title}{Prevalence and measures of
  nutritional compromise among nursing home patients: weight loss, low body
  mass index, malnutrition, and feeding dependency, a systematic review of the
  literature}}.
\newblock {\emph{\JournalTitle{Journal of the American Medical Directors
  Association}}} \textbf{\bibinfo{volume}{14}}, \bibinfo{pages}{94--100}
  (\bibinfo{year}{2013}).

\bibitem{vucea2017}
\bibinfo{author}{Vucea, V.}
\newblock \emph{\bibinfo{title}{Modified Texture Diet and Long Term Care: A
  Secondary Data Analysis of Making the Most of Mealtimes (M3) Project}}.
\newblock Master's thesis, \bibinfo{school}{University of Waterloo}
  (\bibinfo{year}{2017}).

\bibitem{CDC2013}
\bibinfo{author}{Harris-Kojetin, L.}, \bibinfo{author}{Sengupta, M.},
  \bibinfo{author}{Park-Lee, E.} \& \bibinfo{author}{Valverde, R.}
\newblock \bibinfo{journal}{\bibinfo{title}{{Long-term care services in the
  United States: 2013 overview}}}.
\newblock {\emph{\JournalTitle{Vital \& Health Statistics}}}
  \textbf{\bibinfo{volume}{3}} (\bibinfo{year}{2013}).

\bibitem{CIHI_2018}
\bibinfo{author}{{Canadian Institute for Health Information}}.
\newblock \bibinfo{title}{{CCRS Continuing Care Reporting System: Profile} of
  residents in residential and hospital-based continuing care, 2017–2018}.
\newblock
  \bibinfo{howpublished}{\url{https://www.cihi.ca/sites/default/files/document/ccrs-quick-stats-2017-2018-en-web.xlsx}}
  (\bibinfo{year}{2018}).

\bibitem{DoC_2019}
\bibinfo{author}{{Dietitians of Canada}}.
\newblock \bibinfo{title}{{Dietitians of Canada} - resource library: Best
  practices for nutrition, food service and dining in long term care homes}.
\newblock
  \bibinfo{howpublished}{\url{https://www.dietitians.ca/Downloads/Public/2013-Best-Practices-for-Nutrition,-Food-Service-an.aspx}}
  (\bibinfo{year}{2019}).

\bibitem{simmons2000nutritional}
\bibinfo{author}{Simmons, S.~F.} \& \bibinfo{author}{Reuben, D.}
\newblock \bibinfo{journal}{\bibinfo{title}{Nutritional intake monitoring for
  nursing home residents: a comparison of staff documentation, direct
  observation, and photography methods}}.
\newblock {\emph{\JournalTitle{Journal of the American Geriatrics Society}}}
  \textbf{\bibinfo{volume}{48}}, \bibinfo{pages}{209--213}
  (\bibinfo{year}{2000}).

\bibitem{simmons2006feeding}
\bibinfo{author}{Simmons, S.~F.} \& \bibinfo{author}{Schnelle, J.~F.}
\newblock \bibinfo{journal}{\bibinfo{title}{Feeding assistance needs of
  long-stay nursing home residents and staff time to provide care}}.
\newblock {\emph{\JournalTitle{Journal of the American Geriatrics Society}}}
  \textbf{\bibinfo{volume}{54}}, \bibinfo{pages}{919--924}
  (\bibinfo{year}{2006}).

\bibitem{martin2008}
\bibinfo{author}{Martin, C.~K.} \emph{et~al.}
\newblock \bibinfo{journal}{\bibinfo{title}{A novel method to remotely measure
  food intake of free-living individuals in real time: the remote food
  photography method}}.
\newblock {\emph{\JournalTitle{British Journal of Nutrition}}}
  \textbf{\bibinfo{volume}{101}}, \bibinfo{pages}{446--456}
  (\bibinfo{year}{2008}).

\bibitem{williamson2003}
\bibinfo{author}{Williamson, D.~A.} \emph{et~al.}
\newblock \bibinfo{journal}{\bibinfo{title}{Comparison of digital photography
  to weighed and visual estimation of portion sizes}}.
\newblock {\emph{\JournalTitle{Journal of the American Dietetic Association}}}
  \textbf{\bibinfo{volume}{103}}, \bibinfo{pages}{1139--1145}
  (\bibinfo{year}{2003}).

\bibitem{pfisterer2019}
\bibinfo{author}{Pfisterer, K.}, \bibinfo{author}{Boger, J.} \&
  \bibinfo{author}{Wong, A.}
\newblock \bibinfo{journal}{\bibinfo{title}{Prototyping the automated food
  imaging and nutrient intake tracking ({AFINI-T}) system: A modified
  participatory iterative design sprint}}.
\newblock {\emph{\JournalTitle{JMIR Human Factors}}}
  \textbf{\bibinfo{volume}{6}}, \bibinfo{pages}{e13017} (\bibinfo{year}{2019}).

\bibitem{pfisterer2021segmentation}
\bibinfo{author}{Pfisterer, K.~J.} \emph{et~al.}
\newblock \bibinfo{title}{When segmentation is insufficient: Depth-refined
  semantic segmentation rectifies visual-volume discordance in automated
  long-term care food intake tracking} (\bibinfo{year}{2021}).
\newblock \eprint{1910.11250}.

\bibitem{lo2020image}
\bibinfo{author}{Lo, F. P.~W.}, \bibinfo{author}{Sun, Y.},
  \bibinfo{author}{Qiu, J.} \& \bibinfo{author}{Lo, B.}
\newblock \bibinfo{journal}{\bibinfo{title}{Image-based food classification and
  volume estimation for dietary assessment: A review}}.
\newblock {\emph{\JournalTitle{IEEE journal of Biomedical and Health
  Informatics}}} \textbf{\bibinfo{volume}{24}}, \bibinfo{pages}{1926--1939}
  (\bibinfo{year}{2020}).

\bibitem{bruno2017survey}
\bibinfo{author}{Bruno, V.} \& \bibinfo{author}{Silva~Resende, C.~J.}
\newblock \bibinfo{journal}{\bibinfo{title}{A survey on automated food
  monitoring and dietary management systems}}.
\newblock {\emph{\JournalTitle{Journal of Health \& Medical Informatics}}}
  \textbf{\bibinfo{volume}{8}} (\bibinfo{year}{2017}).

\bibitem{doulah2019systematic}
\bibinfo{author}{Doulah, A.}, \bibinfo{author}{Mccrory, M.~A.},
  \bibinfo{author}{Higgins, J.~A.} \& \bibinfo{author}{Sazonov, E.}
\newblock \bibinfo{journal}{\bibinfo{title}{A systematic review of
  technology-driven methodologies for estimation of energy intake}}.
\newblock {\emph{\JournalTitle{IEEE Access}}} \textbf{\bibinfo{volume}{7}},
  \bibinfo{pages}{49653--49668} (\bibinfo{year}{2019}).

\bibitem{pouladzadeh2016}
\bibinfo{author}{Pouladzadeh, P.}, \bibinfo{author}{Shirmohammadi, S.} \&
  \bibinfo{author}{Yassine, A.}
\newblock \bibinfo{journal}{\bibinfo{title}{You are what you eat: So measure
  what you eat!}}
\newblock {\emph{\JournalTitle{IEEE Instrumentation \& Measurement Magazine}}}
  \textbf{\bibinfo{volume}{19}}, \bibinfo{pages}{9--15} (\bibinfo{year}{2016}).

\bibitem{subhi2019}
\bibinfo{author}{Subhi, M.~A.}, \bibinfo{author}{Ali, S.~H.} \&
  \bibinfo{author}{Mohammed, M.~A.}
\newblock \bibinfo{journal}{\bibinfo{title}{Vision-based approaches for
  automatic food recognition and dietary assessment: A survey}}.
\newblock {\emph{\JournalTitle{IEEE Access}}} \textbf{\bibinfo{volume}{7}},
  \bibinfo{pages}{35370--35381} (\bibinfo{year}{2019}).

\bibitem{okamoto2016}
\bibinfo{author}{Okamoto, K.} \& \bibinfo{author}{Yanai, K.}
\newblock \bibinfo{title}{An automatic calorie estimation system of food images
  on a smartphone}.
\newblock In \emph{\bibinfo{booktitle}{Proceedings of the 2nd International
  Workshop on Multimedia Assisted Dietary Management}}, \bibinfo{pages}{63--70}
  (\bibinfo{organization}{ACM}, \bibinfo{year}{2016}).

\bibitem{zhu2015}
\bibinfo{author}{Zhu, F.}, \bibinfo{author}{Bosch, M.},
  \bibinfo{author}{Khanna, N.}, \bibinfo{author}{Boushey, C.~J.} \&
  \bibinfo{author}{Delp, E.~J.}
\newblock \bibinfo{journal}{\bibinfo{title}{Multiple hypotheses image
  segmentation and classification with application to dietary assessment}}.
\newblock {\emph{\JournalTitle{IEEE Journal of Biomedical and Health
  Informatics}}} \textbf{\bibinfo{volume}{19}}, \bibinfo{pages}{377--388}
  (\bibinfo{year}{2015}).

\bibitem{kong2015dietcam}
\bibinfo{author}{Kong, F.}, \bibinfo{author}{He, H.}, \bibinfo{author}{Raynor,
  H.~A.} \& \bibinfo{author}{Tan, J.}
\newblock \bibinfo{journal}{\bibinfo{title}{{DietCam}: multi-view regular shape
  food recognition with a camera phone}}.
\newblock {\emph{\JournalTitle{Pervasive and Mobile Computing}}}
  \textbf{\bibinfo{volume}{19}}, \bibinfo{pages}{108--121}
  (\bibinfo{year}{2015}).

\bibitem{pouladzadeh2014measuring}
\bibinfo{author}{Pouladzadeh, P.}, \bibinfo{author}{Shirmohammadi, S.} \&
  \bibinfo{author}{Al-Maghrabi, R.}
\newblock \bibinfo{journal}{\bibinfo{title}{Measuring calorie and nutrition
  from food image}}.
\newblock {\emph{\JournalTitle{IEEE Transactions on Instrumentation and
  Measurement}}} \textbf{\bibinfo{volume}{63}}, \bibinfo{pages}{1947--1956}
  (\bibinfo{year}{2014}).

\bibitem{meyers2015}
\bibinfo{author}{Meyers, A.} \emph{et~al.}
\newblock \bibinfo{title}{Im2calories: towards an automated mobile vision food
  diary}.
\newblock In \emph{\bibinfo{booktitle}{Proceedings of the IEEE International
  Conference on Computer Vision}}, \bibinfo{pages}{1233--1241}
  (\bibinfo{year}{2015}).

\bibitem{shimoda2015cnn}
\bibinfo{author}{Shimoda, W.} \& \bibinfo{author}{Yanai, K.}
\newblock \bibinfo{title}{{CNN}-based food image segmentation without
  pixel-wise annotation}.
\newblock In \emph{\bibinfo{booktitle}{Proceedings of the International
  Conference on Image Analysis and Processing}}, \bibinfo{pages}{449--457}
  (\bibinfo{year}{2015}).

\bibitem{kawano2015}
\bibinfo{author}{Kawano, Y.} \& \bibinfo{author}{Yanai, K.}
\newblock \bibinfo{journal}{\bibinfo{title}{Foodcam: A real-time food
  recognition system on a smartphone}}.
\newblock {\emph{\JournalTitle{Multimedia Tools and Applications}}}
  \textbf{\bibinfo{volume}{74}}, \bibinfo{pages}{5263--5287}
  (\bibinfo{year}{2015}).

\bibitem{he2013food}
\bibinfo{author}{He, Y.}, \bibinfo{author}{Xu, C.}, \bibinfo{author}{Khanna,
  N.}, \bibinfo{author}{Boushey, C.~J.} \& \bibinfo{author}{Delp, E.~J.}
\newblock \bibinfo{title}{Food image analysis: Segmentation, identification and
  weight estimation}.
\newblock In \emph{\bibinfo{booktitle}{Proceedings of the IEEE International
  Conference on Multimedia and Expo}}, \bibinfo{pages}{1--6}
  (\bibinfo{year}{2013}).

\bibitem{xu2013}
\bibinfo{author}{Xu, C.}, \bibinfo{author}{He, Y.}, \bibinfo{author}{Khanna,
  N.}, \bibinfo{author}{Boushey, C.~J.} \& \bibinfo{author}{Delp, E.~J.}
\newblock \bibinfo{title}{Model-based food volume estimation using 3d pose}.
\newblock In \emph{\bibinfo{booktitle}{Proceedings of the 2013 20th {IEEE}
  {International} {Conference} on Image {Processing} ({ICIP})}},
  \bibinfo{pages}{2534--2538} (\bibinfo{year}{2013}).

\bibitem{chae2011}
\bibinfo{author}{Chae, J.} \emph{et~al.}
\newblock \bibinfo{title}{Volume estimation using food specific shape templates
  in mobile image-based dietary assessment}.
\newblock In \emph{\bibinfo{booktitle}{Proceedings of SPIE}}, vol.
  \bibinfo{volume}{7873}, \bibinfo{pages}{78730K} (\bibinfo{year}{2011}).

\bibitem{jia2014accuracy}
\bibinfo{author}{Jia, W.} \emph{et~al.}
\newblock \bibinfo{journal}{\bibinfo{title}{Accuracy of food portion size
  estimation from digital pictures acquired by a chest-worn camera}}.
\newblock {\emph{\JournalTitle{Public Health Nutrition}}}
  \textbf{\bibinfo{volume}{17}}, \bibinfo{pages}{1671--1681}
  (\bibinfo{year}{2014}).

\bibitem{ofei2019validation}
\bibinfo{author}{Ofei, K.~T.}, \bibinfo{author}{Mikkelsen, B.~E.} \&
  \bibinfo{author}{Scheller, R.~A.}
\newblock \bibinfo{journal}{\bibinfo{title}{Validation of a novel image-weighed
  technique for monitoring food intake and estimation of portion size in
  hospital settings: a pilot study}}.
\newblock {\emph{\JournalTitle{Public Health Nutrition}}}
  \textbf{\bibinfo{volume}{22}}, \bibinfo{pages}{1203--1208}
  (\bibinfo{year}{2019}).

\bibitem{rachakonda2020ilog}
\bibinfo{author}{Rachakonda, L.}, \bibinfo{author}{Mohanty, S.~P.} \&
  \bibinfo{author}{Kougianos, E.}
\newblock \bibinfo{journal}{\bibinfo{title}{{iLog}: an intelligent device for
  automatic food intake monitoring and stress detection in the {IoMT}}}.
\newblock {\emph{\JournalTitle{IEEE Transactions on Consumer Electronics}}}
  \textbf{\bibinfo{volume}{66}}, \bibinfo{pages}{115--124}
  (\bibinfo{year}{2020}).

\bibitem{herzig2020volumetric}
\bibinfo{author}{Herzig, D.} \emph{et~al.}
\newblock \bibinfo{journal}{\bibinfo{title}{Volumetric food quantification
  using computer vision on a depth-sensing smartphone: Preclinical study}}.
\newblock {\emph{\JournalTitle{JMIR mHealth and uHealth}}}
  \textbf{\bibinfo{volume}{8}}, \bibinfo{pages}{e15294} (\bibinfo{year}{2020}).

\bibitem{vucea2019prevalence}
\bibinfo{author}{Vucea, V.} \emph{et~al.}
\newblock \bibinfo{journal}{\bibinfo{title}{Prevalence and characteristics
  associated with modified texture food use in long term care: An analysis of
  making the most of mealtimes (m3) project}}.
\newblock {\emph{\JournalTitle{Canadian Journal of Dietetic Practice and
  Research}}} \textbf{\bibinfo{volume}{80}}, \bibinfo{pages}{104--110}
  (\bibinfo{year}{2019}).

\bibitem{dehais2017}
\bibinfo{author}{Dehais, J.}, \bibinfo{author}{Anthimopoulos, M.},
  \bibinfo{author}{Shevchik, S.} \& \bibinfo{author}{Mougiakakou, S.}
\newblock \bibinfo{journal}{\bibinfo{title}{Two-view 3d {Reconstruction} for
  {Food} {Volume} {Estimation}}}.
\newblock {\emph{\JournalTitle{IEEE Transactions on Multimedia}}}
  \textbf{\bibinfo{volume}{19}}, \bibinfo{pages}{1090--1099}
  (\bibinfo{year}{2017}).

\bibitem{puri2009recognition}
\bibinfo{author}{Puri, M.}, \bibinfo{author}{Zhu, Z.}, \bibinfo{author}{Yu,
  Q.}, \bibinfo{author}{Divakaran, A.} \& \bibinfo{author}{Sawhney, H.}
\newblock \bibinfo{title}{Recognition and volume estimation of food intake
  using a mobile device}.
\newblock In \emph{\bibinfo{booktitle}{2009 Workshop on Applications of
  Computer Vision (WACV)}}, \bibinfo{pages}{1--8} (\bibinfo{year}{2009}).

\bibitem{rahman2012food}
\bibinfo{author}{Rahman, M.~H.} \emph{et~al.}
\newblock \bibinfo{title}{Food volume estimation in a mobile phone based
  dietary assessment system}.
\newblock In \emph{\bibinfo{booktitle}{International Conference on Signal Image
  Technology and Internet Based Systems}}, \bibinfo{pages}{988--995}
  (\bibinfo{year}{2012}).

\bibitem{fang2016}
\bibinfo{author}{Fang, S.} \emph{et~al.}
\newblock \bibinfo{title}{A comparison of food portion size estimation using
  geometric models and depth images}.
\newblock In \emph{\bibinfo{booktitle}{Proceedings of the IEEE International
  Conference on Image Processing (ICIP)}}, \bibinfo{pages}{26--30}
  (\bibinfo{year}{2016}).

\bibitem{shang2011}
\bibinfo{author}{Shang, J.} \emph{et~al.}
\newblock \bibinfo{title}{A mobile structured light system for food volume
  estimation}.
\newblock In \emph{\bibinfo{booktitle}{IEEE International Conference on
  Computer Vision Workshops (ICCV Workshops)}}, \bibinfo{pages}{100--101}
  (\bibinfo{year}{2011}).

\bibitem{chen2012}
\bibinfo{author}{Chen, M.-Y.} \emph{et~al.}
\newblock \bibinfo{title}{Automatic chinese food identification and quantity
  estimation}.
\newblock In \emph{\bibinfo{booktitle}{SIGGRAPH Asia 2012 Technical Briefs}},
  \bibinfo{pages}{29} (\bibinfo{year}{2012}).

\bibitem{liao2016}
\bibinfo{author}{Liao, H.-C.}, \bibinfo{author}{Lim, Z.-Y.} \&
  \bibinfo{author}{Lin, H.-W.}
\newblock \bibinfo{title}{Food intake estimation method using short-range depth
  camera}.
\newblock In \emph{\bibinfo{booktitle}{Signal and Image Processing (ICSIP),
  IEEE International Conference on}}, \bibinfo{pages}{198--204}
  (\bibinfo{year}{2016}).

\bibitem{long2018}
\bibinfo{author}{Long, Y.} \emph{et~al.}
\newblock \bibinfo{journal}{\bibinfo{title}{Potato volume measurement based on
  rgb-d camera}}.
\newblock {\emph{\JournalTitle{IFAC-PapersOnLine}}}
  \textbf{\bibinfo{volume}{51}}, \bibinfo{pages}{515--520}
  (\bibinfo{year}{2018}).

\bibitem{ivorra2014}
\bibinfo{author}{Ivorra, E.}, \bibinfo{author}{Amat, S.~V.},
  \bibinfo{author}{S{\'a}nchez, A.~J.}, \bibinfo{author}{Barat, J.~M.} \&
  \bibinfo{author}{Grau, R.}
\newblock \bibinfo{journal}{\bibinfo{title}{Continuous monitoring of bread
  dough fermentation using a 3d vision structured light technique}}.
\newblock {\emph{\JournalTitle{Journal of Food Engineering}}}
  \textbf{\bibinfo{volume}{130}}, \bibinfo{pages}{8--13}
  (\bibinfo{year}{2014}).

\bibitem{verdu2015}
\bibinfo{author}{Verd{\'u}, S.}, \bibinfo{author}{Ivorra, E.},
  \bibinfo{author}{S{\'a}nchez, A.~J.}, \bibinfo{author}{Barat, J.~M.} \&
  \bibinfo{author}{Grau, R.}
\newblock \bibinfo{journal}{\bibinfo{title}{Relationship between fermentation
  behavior, measured with a 3d vision structured light technique, and the
  internal structure of bread}}.
\newblock {\emph{\JournalTitle{Journal of Food Engineering}}}
  \textbf{\bibinfo{volume}{146}}, \bibinfo{pages}{227--233}
  (\bibinfo{year}{2015}).

\bibitem{perez2017}
\bibinfo{author}{Perez, R.~M.}, \bibinfo{author}{Cheein, F.~A.} \&
  \bibinfo{author}{Rosell-Polo, J.~R.}
\newblock \bibinfo{journal}{\bibinfo{title}{Flexible system of multiple rgb-d
  sensors for measuring and classifying fruits in agri-food industry}}.
\newblock {\emph{\JournalTitle{Computers and Electronics in Agriculture}}}
  \textbf{\bibinfo{volume}{139}}, \bibinfo{pages}{231--242}
  (\bibinfo{year}{2017}).

\bibitem{lu2017}
\bibinfo{author}{Lu, Y.} \& \bibinfo{author}{Lu, R.}
\newblock \bibinfo{journal}{\bibinfo{title}{Using composite sinusoidal patterns
  in structured-illumination reflectance imaging (siri) for enhanced detection
  of apple bruise}}.
\newblock {\emph{\JournalTitle{Journal of Food Engineering}}}
  \textbf{\bibinfo{volume}{199}}, \bibinfo{pages}{54--64}
  (\bibinfo{year}{2017}).

\bibitem{lu2018}
\bibinfo{author}{Lu, Y.} \& \bibinfo{author}{Lu, R.}
\newblock \bibinfo{journal}{\bibinfo{title}{Structured-illumination reflectance
  imaging coupled with phase analysis techniques for surface profiling of
  apples}}.
\newblock {\emph{\JournalTitle{Journal of Food Engineering}}}
  \textbf{\bibinfo{volume}{232}}, \bibinfo{pages}{11--20}
  (\bibinfo{year}{2018}).

\bibitem{shang2012}
\bibinfo{author}{Shang, J.} \emph{et~al.}
\newblock \bibinfo{title}{Dietary intake assessment using integrated sensors
  and software}.
\newblock In \emph{\bibinfo{booktitle}{Multimedia on Mobile Devices 2012; and
  Multimedia Content Access: Algorithms and Systems VI}}, vol.
  \bibinfo{volume}{8304}, \bibinfo{pages}{830403}
  (\bibinfo{organization}{International Society for Optics and Photonics},
  \bibinfo{year}{2012}).

\bibitem{castellanos2002}
\bibinfo{author}{Castellanos, V.~H.} \& \bibinfo{author}{Andrews, Y.~N.}
\newblock \bibinfo{journal}{\bibinfo{title}{Inherent flaws in a method of
  estimating meal intake commonly used in long-term-care facilities}}.
\newblock {\emph{\JournalTitle{Journal of the American Dietetic Association}}}
  \textbf{\bibinfo{volume}{102}}, \bibinfo{pages}{826--830}
  (\bibinfo{year}{2002}).

\bibitem{zhou2019application}
\bibinfo{author}{Zhou, L.}, \bibinfo{author}{Zhang, C.}, \bibinfo{author}{Liu,
  F.}, \bibinfo{author}{Qiu, Z.} \& \bibinfo{author}{He, Y.}
\newblock \bibinfo{journal}{\bibinfo{title}{Application of deep learning in
  food: a review}}.
\newblock {\emph{\JournalTitle{Comprehensive Reviews in Food Science and Food
  Safety}}} \textbf{\bibinfo{volume}{18}}, \bibinfo{pages}{1793--1811}
  (\bibinfo{year}{2019}).

\bibitem{cioccaJBHI}
\bibinfo{author}{Ciocca, G.}, \bibinfo{author}{Napoletano, P.} \&
  \bibinfo{author}{Schettini, R.}
\newblock \bibinfo{journal}{\bibinfo{title}{Food recognition: a new dataset,
  experiments and results}}.
\newblock {\emph{\JournalTitle{IEEE Journal of Biomedical and Health
  Informatics}}} \textbf{\bibinfo{volume}{21}}, \bibinfo{pages}{588--598}
  (\bibinfo{year}{2017}).

\bibitem{lo2018food}
\bibinfo{author}{Lo, F. P.-W.}, \bibinfo{author}{Sun, Y.},
  \bibinfo{author}{Qiu, J.} \& \bibinfo{author}{Lo, B.}
\newblock \bibinfo{journal}{\bibinfo{title}{Food volume estimation based on
  deep learning view synthesis from a single depth map}}.
\newblock {\emph{\JournalTitle{Nutrients}}} \textbf{\bibinfo{volume}{10}},
  \bibinfo{pages}{2005} (\bibinfo{year}{2018}).

\bibitem{lo2019point2volume}
\bibinfo{author}{Lo, F. P.-W.}, \bibinfo{author}{Sun, Y.},
  \bibinfo{author}{Qiu, J.} \& \bibinfo{author}{Lo, B.~P.}
\newblock \bibinfo{journal}{\bibinfo{title}{Point2volume: A vision-based
  dietary assessment approach using view synthesis}}.
\newblock {\emph{\JournalTitle{IEEE Transactions on Industrial Informatics}}}
  \textbf{\bibinfo{volume}{16}}, \bibinfo{pages}{577--586}
  (\bibinfo{year}{2019}).

\bibitem{meyers2015im2calories}
\bibinfo{author}{Meyers, A.} \emph{et~al.}
\newblock \bibinfo{title}{Im2calories: towards an automated mobile vision food
  diary}.
\newblock In \emph{\bibinfo{booktitle}{Proceedings of the IEEE International
  Conference on Computer Vision}}, \bibinfo{pages}{1233--1241}
  (\bibinfo{year}{2015}).

\bibitem{aslan2018semantic}
\bibinfo{author}{Aslan, S.}, \bibinfo{author}{Ciocca, G.} \&
  \bibinfo{author}{Schettini, R.}
\newblock \bibinfo{title}{Semantic food segmentation for automatic dietary
  monitoring}.
\newblock In \emph{\bibinfo{booktitle}{Proceedings of the IEEE International
  Conference on Consumer Electronics-Berlin}}, \bibinfo{pages}{1--6}
  (\bibinfo{year}{2018}).

\bibitem{aguilar2018grab}
\bibinfo{author}{Aguilar, E.}, \bibinfo{author}{Remeseiro, B.},
  \bibinfo{author}{Bola{\~n}os, M.} \& \bibinfo{author}{Radeva, P.}
\newblock \bibinfo{journal}{\bibinfo{title}{Grab, pay, and eat: Semantic food
  detection for smart restaurants}}.
\newblock {\emph{\JournalTitle{IEEE Transactions on Multimedia}}}
  \textbf{\bibinfo{volume}{20}}, \bibinfo{pages}{3266--3275}
  (\bibinfo{year}{2018}).

\bibitem{boushey2017}
\bibinfo{author}{Boushey, C.~J.}, \bibinfo{author}{Spoden, M.},
  \bibinfo{author}{Zhu, F.~M.}, \bibinfo{author}{Delp, E.~J.} \&
  \bibinfo{author}{Kerr, D.~A.}
\newblock \bibinfo{journal}{\bibinfo{title}{New mobile methods for dietary
  assessment: review of image-assisted and image-based dietary assessment
  methods}}.
\newblock {\emph{\JournalTitle{Proceedings of the Nutrition Society}}}
  \textbf{\bibinfo{volume}{76}}, \bibinfo{pages}{283–294},
  \doiprefix\url{10.1017/S0029665116002913} (\bibinfo{year}{2017}).

\bibitem{selamat2020automatic}
\bibinfo{author}{Selamat, N.~A.} \& \bibinfo{author}{Ali, S. H.~M.}
\newblock \bibinfo{journal}{\bibinfo{title}{Automatic food intake monitoring
  based on chewing activity: A survey}}.
\newblock {\emph{\JournalTitle{IEEE Access}}} \textbf{\bibinfo{volume}{8}},
  \bibinfo{pages}{48846--48869} (\bibinfo{year}{2020}).

\bibitem{moguel2019systematic}
\bibinfo{author}{Moguel, E.}, \bibinfo{author}{Berrocal, J.} \&
  \bibinfo{author}{Garc{\'\i}a-Alonso, J.}
\newblock \bibinfo{journal}{\bibinfo{title}{Systematic literature review of
  food-intake monitoring in an aging population}}.
\newblock {\emph{\JournalTitle{Sensors}}} \textbf{\bibinfo{volume}{19}},
  \bibinfo{pages}{3265} (\bibinfo{year}{2019}).

\bibitem{vu2017wearable}
\bibinfo{author}{Vu, T.}, \bibinfo{author}{Lin, F.},
  \bibinfo{author}{Alshurafa, N.} \& \bibinfo{author}{Xu, W.}
\newblock \bibinfo{journal}{\bibinfo{title}{Wearable food intake monitoring
  technologies: A comprehensive review}}.
\newblock {\emph{\JournalTitle{Computers}}} \textbf{\bibinfo{volume}{6}},
  \bibinfo{pages}{4} (\bibinfo{year}{2017}).

\bibitem{pettitt2016pilot}
\bibinfo{author}{Pettitt, C.} \emph{et~al.}
\newblock \bibinfo{journal}{\bibinfo{title}{A pilot study to determine whether
  using a lightweight, wearable micro-camera improves dietary assessment
  accuracy and offers information on macronutrients and eating rate}}.
\newblock {\emph{\JournalTitle{British Journal of Nutrition}}}
  \textbf{\bibinfo{volume}{115}}, \bibinfo{pages}{160--167}
  (\bibinfo{year}{2016}).

\bibitem{beltran2016adapting}
\bibinfo{author}{Beltran, A.} \emph{et~al.}
\newblock \bibinfo{title}{Adapting the ebutton to the abilities of children for
  diet assessment}.
\newblock In \emph{\bibinfo{booktitle}{Proceedings of Measuring Behavior 2016:
  10th International Conference on Methods and Techniques in Behavioral
  Research. International Conference on Methods and Techniques in Behavioral
  Research (10th: 2016: Dublin, Ireland)}}, vol. \bibinfo{volume}{2016},
  \bibinfo{pages}{72} (\bibinfo{organization}{NIH Public Access},
  \bibinfo{year}{2016}).

\bibitem{pouladzadeh2016food}
\bibinfo{author}{Pouladzadeh, P.}, \bibinfo{author}{Kuhad, P.},
  \bibinfo{author}{Peddi, S. V.~B.}, \bibinfo{author}{Yassine, A.} \&
  \bibinfo{author}{Shirmohammadi, S.}
\newblock \bibinfo{title}{Food calorie measurement using deep learning neural
  network}.
\newblock In \emph{\bibinfo{booktitle}{Proceedings of the IEEE International
  Instrumentation and Measurement Technology}}, \bibinfo{pages}{1--6}
  (\bibinfo{year}{2016}).

\bibitem{wang2018}
\bibinfo{author}{Wang, Y.}, \bibinfo{author}{He, Y.}, \bibinfo{author}{Boushey,
  C.~J.}, \bibinfo{author}{Zhu, F.} \& \bibinfo{author}{Delp, E.~J.}
\newblock \bibinfo{journal}{\bibinfo{title}{Context based image analysis with
  application in dietary assessment and evaluation}}.
\newblock {\emph{\JournalTitle{Multimedia Tools and Applications}}}
  \textbf{\bibinfo{volume}{77}}, \bibinfo{pages}{19769--19794}
  (\bibinfo{year}{2018}).

\bibitem{yunus2018framework}
\bibinfo{author}{Yunus, R.} \emph{et~al.}
\newblock \bibinfo{journal}{\bibinfo{title}{A framework to estimate the
  nutritional value of food in real time using deep learning techniques}}.
\newblock {\emph{\JournalTitle{IEEE Access}}} \textbf{\bibinfo{volume}{7}},
  \bibinfo{pages}{2643--2652} (\bibinfo{year}{2018}).

\bibitem{hassannejad2017}
\bibinfo{author}{Hassannejad, H.} \emph{et~al.}
\newblock \bibinfo{journal}{\bibinfo{title}{A new approach to image-based
  estimation of food volume}}.
\newblock {\emph{\JournalTitle{Algorithms}}} \textbf{\bibinfo{volume}{10}},
  \bibinfo{pages}{66} (\bibinfo{year}{2017}).

\bibitem{parent2012}
\bibinfo{author}{Parent, M.}, \bibinfo{author}{Niezgoda, H.},
  \bibinfo{author}{Keller, H.~H.}, \bibinfo{author}{Chambers, L.~W.} \&
  \bibinfo{author}{Daly, S.}
\newblock \bibinfo{journal}{\bibinfo{title}{Comparison of visual estimation
  methods for regular and modified textures: real-time vs digital imaging}}.
\newblock {\emph{\JournalTitle{Journal of the Academy of Nutrition and
  Dietetics}}} \textbf{\bibinfo{volume}{112}}, \bibinfo{pages}{1636--1641}
  (\bibinfo{year}{2012}).

\bibitem{astell2014validation}
\bibinfo{author}{Astell, A.~J.} \emph{et~al.}
\newblock \bibinfo{journal}{\bibinfo{title}{Validation of the nana (novel
  assessment of nutrition and ageing) touch screen system for use at home by
  older adults}}.
\newblock {\emph{\JournalTitle{Experimental Gerontology}}}
  \textbf{\bibinfo{volume}{60}}, \bibinfo{pages}{100--107}
  (\bibinfo{year}{2014}).

\bibitem{huskisson2007}
\bibinfo{author}{Huskisson, E.}, \bibinfo{author}{Maggini, S.} \&
  \bibinfo{author}{Ruf, M.}
\newblock \bibinfo{journal}{\bibinfo{title}{The influence of micronutrients on
  cognitive function and performance}}.
\newblock {\emph{\JournalTitle{{Journal of International Medical Research}}}}
  \textbf{\bibinfo{volume}{35}}, \bibinfo{pages}{1--19} (\bibinfo{year}{2007}).

\bibitem{reginaldo2014}
\bibinfo{author}{Reginaldo, C.} \emph{et~al.}
\newblock \bibinfo{journal}{\bibinfo{title}{The association between vitamin
  {B6} and cognitive decline is modified by inflammatory state}}.
\newblock {\emph{\JournalTitle{{The FASEB Journal}}}}
  \textbf{\bibinfo{volume}{28}}, \bibinfo{pages}{LB425} (\bibinfo{year}{2014}).

\bibitem{keller2018prevalence}
\bibinfo{author}{Keller, H.~H.} \emph{et~al.}
\newblock \bibinfo{journal}{\bibinfo{title}{Prevalence of inadequate
  micronutrient intakes of {C}anadian long-term care residents}}.
\newblock {\emph{\JournalTitle{British Journal of Nutrition}}}
  \textbf{\bibinfo{volume}{119}}, \bibinfo{pages}{1047--1056}
  (\bibinfo{year}{2018}).

\bibitem{CFG2019}
\bibinfo{author}{{Government of Canada}}.
\newblock \bibinfo{title}{{Canada's Food Guide}}.
\newblock \bibinfo{howpublished}{\url{https://food-guide.canada.ca/en/}}
  (\bibinfo{year}{(accessed: 24.08.2021)}).

\bibitem{pouladzadeh2014}
\bibinfo{author}{Pouladzadeh, P.}, \bibinfo{author}{Shirmohammadi, S.} \&
  \bibinfo{author}{Yassine, A.}
\newblock \bibinfo{title}{Using graph cut segmentation for food calorie
  measurement}.
\newblock In \emph{\bibinfo{booktitle}{Proceedings of the IEEE International
  Symposium on Medical Measurements and Applications}}, \bibinfo{pages}{1--6}
  (\bibinfo{year}{2014}).

\bibitem{zhang2012}
\bibinfo{author}{Zhang, Y.} \& \bibinfo{author}{Wu, L.}
\newblock \bibinfo{journal}{\bibinfo{title}{Classification of fruits using
  computer vision and a multiclass support vector machine}}.
\newblock {\emph{\JournalTitle{Sensors}}} \textbf{\bibinfo{volume}{12}},
  \bibinfo{pages}{12489--12505} (\bibinfo{year}{2012}).

\bibitem{bolle1996veggievision}
\bibinfo{author}{Bolle, R.~M.}, \bibinfo{author}{Connell, J.~H.},
  \bibinfo{author}{Haas, N.}, \bibinfo{author}{Mohan, R.} \&
  \bibinfo{author}{Taubin, G.}
\newblock \bibinfo{title}{Veggievision: A produce recognition system}.
\newblock In \emph{\bibinfo{booktitle}{Proceedings Third IEEE Workshop on
  Applications of Computer Vision. WACV'96}}, \bibinfo{pages}{244--251}
  (\bibinfo{organization}{IEEE}, \bibinfo{year}{1996}).

\bibitem{rocha2010automatic}
\bibinfo{author}{Rocha, A.}, \bibinfo{author}{Hauagge, D.~C.},
  \bibinfo{author}{Wainer, J.} \& \bibinfo{author}{Goldenstein, S.}
\newblock \bibinfo{journal}{\bibinfo{title}{Automatic fruit and vegetable
  classification from images}}.
\newblock {\emph{\JournalTitle{Computers and Electronics in Agriculture}}}
  \textbf{\bibinfo{volume}{70}}, \bibinfo{pages}{96--104}
  (\bibinfo{year}{2010}).

\bibitem{arivazhagan2010fruit}
\bibinfo{author}{Arivazhagan, S.}, \bibinfo{author}{Shebiah, R.~N.},
  \bibinfo{author}{Nidhyanandhan, S.~S.} \& \bibinfo{author}{Ganesan, L.}
\newblock \bibinfo{journal}{\bibinfo{title}{Fruit recognition using color and
  texture features}}.
\newblock {\emph{\JournalTitle{Journal of Emerging Trends in Computing and
  Information Sciences}}} \textbf{\bibinfo{volume}{1}}, \bibinfo{pages}{90--94}
  (\bibinfo{year}{2010}).

\bibitem{chowdhury2013vegetables}
\bibinfo{author}{Chowdhury, M.~T.}, \bibinfo{author}{Alam, M.~S.},
  \bibinfo{author}{Hasan, M.~A.} \& \bibinfo{author}{Khan, M.~I.}
\newblock \bibinfo{journal}{\bibinfo{title}{Vegetables detection from the
  glossary shop for the blind}}.
\newblock {\emph{\JournalTitle{IOSR Journal of Electrical and Electronics
  Engineering}}} \textbf{\bibinfo{volume}{8}}, \bibinfo{pages}{43--53}
  (\bibinfo{year}{2013}).

\bibitem{chen2012automatic}
\bibinfo{author}{Chen, M.-Y.} \emph{et~al.}
\newblock \bibinfo{title}{Automatic chinese food identification and quantity
  estimation}.
\newblock In \emph{\bibinfo{booktitle}{SIGGRAPH Asia 2012 Technical Briefs}},
  \bibinfo{pages}{1--4} (\bibinfo{year}{2012}).

\bibitem{miyazaki2011}
\bibinfo{author}{Miyazaki, T.}, \bibinfo{author}{de~Silva, G.~C.} \&
  \bibinfo{author}{Aizawa, K.}
\newblock \bibinfo{title}{Image-based calorie content estimation for dietary
  assessment}.
\newblock In \emph{\bibinfo{booktitle}{Multimedia (ISM), 2011 IEEE
  International Symposium on}}, \bibinfo{pages}{363--368}
  (\bibinfo{organization}{IEEE}, \bibinfo{year}{2011}).

\bibitem{ege2017}
\bibinfo{author}{Ege, T.} \& \bibinfo{author}{Yanai, K.}
\newblock \bibinfo{title}{Simultaneous estimation of food categories and
  calories with multi-task cnn}.
\newblock In \emph{\bibinfo{booktitle}{Machine Vision Applications (MVA), 2017
  Fifteenth IAPR International Conference on}}, \bibinfo{pages}{198--201}
  (\bibinfo{organization}{IEEE}, \bibinfo{year}{2017}).

\bibitem{fang2015}
\bibinfo{author}{Fang, S.}, \bibinfo{author}{Liu, C.}, \bibinfo{author}{Zhu,
  F.}, \bibinfo{author}{Delp, E.~J.} \& \bibinfo{author}{Boushey, C.~J.}
\newblock \bibinfo{title}{Single-view food portion estimation based on
  geometric models}.
\newblock In \emph{\bibinfo{booktitle}{Multimedia (ISM), 2015 IEEE
  International Symposium on}}, \bibinfo{pages}{385--390}
  (\bibinfo{organization}{IEEE}, \bibinfo{year}{2015}).

\bibitem{chokr2017}
\bibinfo{author}{Chokr, M.} \& \bibinfo{author}{Elbassuoni, S.}
\newblock \bibinfo{title}{Calories prediction from food images.}
\newblock In \emph{\bibinfo{booktitle}{AAAI}}, \bibinfo{pages}{4664--4669}
  (\bibinfo{year}{2017}).

\bibitem{ispirova2020evaluating}
\bibinfo{author}{Ispirova, G.}, \bibinfo{author}{Eftimov, T.} \&
  \bibinfo{author}{Seljak, B.~K.}
\newblock \bibinfo{journal}{\bibinfo{title}{Evaluating missing value imputation
  methods for food composition databases}}.
\newblock {\emph{\JournalTitle{Food and Chemical Toxicology}}}
  \textbf{\bibinfo{volume}{141}}, \bibinfo{pages}{111368}
  (\bibinfo{year}{2020}).

\bibitem{holden2008assessing}
\bibinfo{author}{Holden, J.~M.} \& \bibinfo{author}{Lemar, L.~E.}
\newblock \bibinfo{journal}{\bibinfo{title}{Assessing vitamin d contents in
  foods and supplements: challenges and needs}}.
\newblock {\emph{\JournalTitle{{The American Journal of Clinical Nutrition}}}}
  \textbf{\bibinfo{volume}{88}}, \bibinfo{pages}{551S--553S}
  (\bibinfo{year}{2008}).

\bibitem{xia2015situ}
\bibinfo{author}{Xia, C.}, \bibinfo{author}{Wang, L.}, \bibinfo{author}{Chung,
  B.-K.} \& \bibinfo{author}{Lee, J.-M.}
\newblock \bibinfo{journal}{\bibinfo{title}{In situ 3d segmentation of
  individual plant leaves using a rgb-d camera for agricultural automation}}.
\newblock {\emph{\JournalTitle{Sensors}}} \textbf{\bibinfo{volume}{15}},
  \bibinfo{pages}{20463--20479} (\bibinfo{year}{2015}).

\bibitem{lin2020color}
\bibinfo{author}{Lin, G.}, \bibinfo{author}{Tang, Y.}, \bibinfo{author}{Zou,
  X.}, \bibinfo{author}{Xiong, J.} \& \bibinfo{author}{Fang, Y.}
\newblock \bibinfo{journal}{\bibinfo{title}{Color-, depth-, and shape-based 3d
  fruit detection}}.
\newblock {\emph{\JournalTitle{Precision Agriculture}}}
  \textbf{\bibinfo{volume}{21}}, \bibinfo{pages}{1--17} (\bibinfo{year}{2020}).

\bibitem{gai2020automated}
\bibinfo{author}{Gai, J.}, \bibinfo{author}{Tang, L.} \&
  \bibinfo{author}{Steward, B.~L.}
\newblock \bibinfo{journal}{\bibinfo{title}{Automated crop plant detection
  based on the fusion of color and depth images for robotic weed control}}.
\newblock {\emph{\JournalTitle{Journal of Field Robotics}}}
  \textbf{\bibinfo{volume}{37}}, \bibinfo{pages}{35--52}
  (\bibinfo{year}{2020}).

\bibitem{beckman2019deep}
\bibinfo{author}{Beckman, G.~H.}, \bibinfo{author}{Polyzois, D.} \&
  \bibinfo{author}{Cha, Y.-J.}
\newblock \bibinfo{journal}{\bibinfo{title}{Deep learning-based automatic
  volumetric damage quantification using depth camera}}.
\newblock {\emph{\JournalTitle{Automation in Construction}}}
  \textbf{\bibinfo{volume}{99}}, \bibinfo{pages}{114--124}
  (\bibinfo{year}{2019}).

\bibitem{danielczuk2019segmenting}
\bibinfo{author}{Danielczuk, M.} \emph{et~al.}
\newblock \bibinfo{title}{Segmenting unknown 3d objects from real depth images
  using mask r-cnn trained on synthetic data}.
\newblock In \emph{\bibinfo{booktitle}{2019 International Conference on
  Robotics and Automation (ICRA)}}, \bibinfo{pages}{7283--7290}
  (\bibinfo{organization}{IEEE}, \bibinfo{year}{2019}).

\bibitem{OLTCA2020}
\bibinfo{author}{Association, O. L.-T.~C.}
\newblock \bibinfo{title}{The role of long-term care} (\bibinfo{year}{2020}).

\bibitem{duizer2020menuplanning}
\bibinfo{author}{Duizer, L.~M.} \& \bibinfo{author}{Keller, H.~H.}
\newblock \bibinfo{journal}{\bibinfo{title}{Planning micronutrient-dense menus
  in ontario long-term care homes: Strategies and challenges}}.
\newblock {\emph{\JournalTitle{Canadian Journal of Dietetic Practice and
  Research}}} \textbf{\bibinfo{volume}{81}}, \bibinfo{pages}{198--203},
  \doiprefix\url{10.3148/cjdpr-2020-014} (\bibinfo{year}{2020}).
\newblock \bibinfo{note}{PMID: 32495638},
  \eprint{https://doi.org/10.3148/cjdpr-2020-014}.

\bibitem{grieger2007nutrient}
\bibinfo{author}{Grieger, J.} \& \bibinfo{author}{Nowson, C.}
\newblock \bibinfo{journal}{\bibinfo{title}{Nutrient intake and plate waste
  from an australian residential care facility}}.
\newblock {\emph{\JournalTitle{European journal of clinical nutrition}}}
  \textbf{\bibinfo{volume}{61}}, \bibinfo{pages}{655--663}
  (\bibinfo{year}{2007}).

\bibitem{chun2019grounded}
\bibinfo{author}{Chun~Tie, Y.}, \bibinfo{author}{Birks, M.} \&
  \bibinfo{author}{Francis, K.}
\newblock \bibinfo{journal}{\bibinfo{title}{Grounded theory research: A design
  framework for novice researchers}}.
\newblock {\emph{\JournalTitle{SAGE open medicine}}}
  \textbf{\bibinfo{volume}{7}}, \bibinfo{pages}{2050312118822927}
  (\bibinfo{year}{2019}).

\bibitem{kaur2019foodx}
\bibinfo{author}{Kaur, P.} \emph{et~al.}
\newblock \bibinfo{journal}{\bibinfo{title}{Foodx-251: A dataset for
  fine-grained food classification}}.
\newblock {\emph{\JournalTitle{arXiv preprint arXiv:1907.06167}}}
  (\bibinfo{year}{2019}).

\bibitem{pfisterer2018}
\bibinfo{author}{Pfisterer, K.~J.}, \bibinfo{author}{Amelard, R.},
  \bibinfo{author}{Chung, A.~G.} \& \bibinfo{author}{Wong, A.}
\newblock \bibinfo{journal}{\bibinfo{title}{A new take on measuring relative
  nutritional density: The feasibility of using a deep neural network to assess
  commercially-prepared pur{\'e}ed food concentrations}}.
\newblock {\emph{\JournalTitle{Journal of Food Engineering}}}
  \textbf{\bibinfo{volume}{223}}, \bibinfo{pages}{220--235}
  (\bibinfo{year}{2018}).

\bibitem{simonyan2014}
\bibinfo{author}{Simonyan, K.} \& \bibinfo{author}{Zisserman, A.}
\newblock \bibinfo{journal}{\bibinfo{title}{Very deep convolutional networks
  for large-scale image recognition}}.
\newblock {\emph{\JournalTitle{arXiv preprint arXiv:1409.1556}}}
  (\bibinfo{year}{2014}).

\bibitem{giavarina2015}
\bibinfo{author}{Giavarina, D.}
\newblock \bibinfo{journal}{\bibinfo{title}{Understanding {Bland Altman}
  analysis}}.
\newblock {\emph{\JournalTitle{Biochemia Medica}}}
  \textbf{\bibinfo{volume}{25}}, \bibinfo{pages}{141--151}
  (\bibinfo{year}{2015}).

\end{thebibliography}

\section*{Acknowledgements}

This work was funded by the National Science and Engineering Research Council of Canada (NSERC PGSD3-489611-2016, NSERC PDF-503038-2017) , the National Institute for Health Research (CIHR 202012MFE-459277), and the Canada Research Chairs (CRC) program.

\section*{Author contributions statement}

K.J.P and R.A contributed equally to this work. K.J.P conceptualized the system; R.A and A.W provided additional contributions to system design. K.J.P was the main contributor to experimental design, and contributed to algorithmic design. R.A was the main contributor to algorithmic design with additional contributions from K.J.P, A.W. A.G.C provided additional support on data collection and preliminary analyses during the initial conceptualization of this project. K.J.P was the main contributor to data analyses; K.J.P and R.A conducted data analyses. H.H.K provided clinical nutrition direction and perspective. J.B oversaw the user-study elements. A.W oversaw the project as a whole. K.J.P was the main contributor to writing the manuscript with additional contributions from R.A. All authors reviewed the manuscript.

\end{document}


\maketitle

\section*{S1. Foods included in dataset}
\label{S1:includedfoods}
Table~\ref{tab:foodsimaged} provides an overview of all foods included as part of the regular texture and modified texture food datasets included.
\begin{sidewaystable}[]
\caption[ Cumulative imaged foods list.]{Cumulative list of foods imaged.}
\centering
\label{tab:foodsimaged}
\begin{tabular}{llll}
\textbf{Food   Component} &
  \textbf{Regular Texture foods} &
  \textbf{\begin{tabular}[c]{@{}l@{}}Modified Texture Foods\\ (with recipes as outined in~\cite{pfisterer2021segmentation})\end{tabular}} &
  \textbf{\begin{tabular}[c]{@{}l@{}}Additional Modified Texture Foods\\ (with segmentations)\end{tabular}} \\ \hline
\textit{\textbf{Grains}} &
  Cheese tortellini /w tomato sauce &
  Bow tie pasta /w carbonara sauce &
  Basmati Rice \\
                               & Oatmeal            & Macaroni salad                &                                             \\
                               & Whole wheat toast & Vegetable rotini              &                                             \\ \hline
 \textit{\textbf{Vegetables and fruits}}  & Corn               & Asian vegetables              & Beet \& onion salad                          \\ 
 &
  Mashed potatoes &
  Baked polenta /w garlic &
  Cantaloupe Chunks \\
                               & Mixed greens salad & California vegetables         & Green beans with pimento                    \\
                               &                    & Greek salad                   & Grilled vegetable salad                     \\
                               &                    & Mango \& pineapple            & Roasted cauliflower                         \\
                               &                    & Red potato salad              &                                             \\
                               &                    & Sauteed spinach \& kale       &                                             \\
                               &                    & Seasoned green peas           &                                             \\
                               &                    & Stewed rhubarb \& berries    &                                             \\
                               &                    & Strawberries \& bananas       &                                             \\
                               &                    & Sweet and sour cabbage        &                                             \\ \hline
\textit{\textbf{Proteins}}     & Meatloaf           & Baked basa                    & Bean \& sausage strata                       \\
                               & Scrambled egg      & Braised beef liver \& onions  & Grilled lemon \& garlic chicken             \\ 
                               &                    & Braised lamb shanks           & Pork tortiere                               \\
                               &                    & Hot dog wiener                & Roast beef with miracle whip                \\
                               &                    & Orange ginger chicken         &                                             \\
                               &                    & Salisbury steak \& gravy      &                                             \\
                               &                    & Teriyaki meatballs            &                                             \\
                               &                    & Tuna salad                    &                                             \\ \hline
\textit{\textbf{Mixed dishes}} & Oatmeal cookie     & Barley beef soup              & Black bean soup                             \\
                               &                    & Blueberry coffee crumble cake & Broken glass parfait                        \\
                               &                    & Eggplant parmigiana           & Butternut squash soup                       \\ 
                               &                    & English trifle                & Cranberry spice oatmeal cookie              \\
                               &                    & Lemon chicken orzo soup       & Lemon meringue pie                          \\
                               &                    &                               & Peach jello                                 \\
                               &                    &                               & Pear crumble cake                           \\
                               &                    &                               & Roast beef with miracle whip on whole wheat \\
                               &                    &                               & Turkey burger on wheat bun                  \\ \hline
\end{tabular}
\end{sidewaystable}

\section*{S2. Nutrient Intake Accuracies}
Tables~\ref{tab:nut_est_accuracy_macros},~\ref{tab:nut_est_accuracy_elements},and~\ref{tab:nut_est_accuracy_vitamins} provide a comprehensive overview of nutrient intake accuracy within and across our LTC datasets. For each of the following nutrients of interest, Figures~\ref{fig:calories} to~\ref{fig:vitK} show the correlation and agreement between mass and volume estimates methods for determining nutritional intake at the whole plate level across all imaged samples. The left panel depicts the goodness of fit with linear regression and coefficient of determination ($r^2$), right depicts the degree of agreement between measures and bias from the Bland-Altman method.

\begin{table}[]
\caption [Macronutrient intake accuracies within and across datasets.]{Macronutrient intake accuracies within and across datasets.}
\label{tab:nut_est_accuracy_macros}
\small
\centering
\begin{tabular}{llccccc}
\hline
\multicolumn{2}{l}{\textbf{Dataset}} & \multicolumn{5}{c}{\textbf{\begin{tabular}[c]{@{}c@{}}Nutrient intake   accuracy\\      $\mu \pm \sigma$ (\% error)\end{tabular}}} \\
\textbf{Meal} & \textbf{\begin{tabular}[c]{@{}l@{}}\# of classes\\ (\# images)\end{tabular}} & \textbf{Calories (kcal)} & \textbf{Carbohydrates (g)} & \textbf{Fibre (g)} & \textbf{Fats (g)} & \textbf{Protein (g)} \\ \hline
RTD: B & 3 (125) & 18.13 $\pm$ 16.76 (18) & 2.68 $\pm$ 2.36 (3) & 0.50 $\pm$ 0.48 (1) & 0.67 $\pm$ 0.74 (1) & 1.08 $\pm$ 1.07 (1) \\
RTD: L & 3 (125) & 23.43 $\pm$ 21.64 (23) & 2.70 $\pm$ 2.99 (3) & 0.47 $\pm$ 0.30 (0) & 0.88 $\pm$ 0.82 (1) & 0.82 $\pm$ 0.85 (1) \\
RTD: D & 3 (125) & 17.04 $\pm$ 17.38 (17) & 2.57 $\pm$ 2.37 (3) & 0.22 $\pm$ 0.23 (0) & 0.66 $\pm$ 0.80 (1) & 0.59 $\pm$ 0.75 (1) \\ \hline
\textit{        RTF subtotal} & \textit{9 (375)} & \textit{16.31 $\pm$ 13.63 (16)} & \textit{2.11 $\pm$ 2.05 (2)} & \textit{0.37 $\pm$ 0.39 (0)} & \textit{0.83 $\pm$ 0.80 (1)} & \textit{1.11 $\pm$ 0.98 (1)} \\ \hline
MTD: D1 - L & 15 (90) & 9.31 $\pm$ 13.27 (9) & 1.38 $\pm$ 1.52 (1) & 0.08 $\pm$ 0.07 (0) & 0.28 $\pm$ 0.57 (0) & 0.31 $\pm$ 0.52 (0) \\
MTD: D1 - D & 5 (25) & 28.17 $\pm$ 25.01 (28) & 1.33 $\pm$ 1.06 (1) & 0.09 $\pm$ 0.07 (0) & 1.19 $\pm$ 1.09 (1) & 2.93 $\pm$ 2.69 (3) \\
MTD: D2- L & 12 (74) & n/a & n/a & n/a & n/a & n/a \\
MTD: D2 - D & 12 (91) & 37.25 $\pm$ 38.66 (37) & 3.26 $\pm$ 4.10 (3) & 0.11 $\pm$ 0.14 (0) & 1.92 $\pm$ 1.83 (2) & 1.77 $\pm$ 1.57 (2) \\
MTD: D3 - L & 10 (85) & 12.46 $\pm$ 9.61 (12) & 1.41 $\pm$ 1.15 (1) & 0.13 $\pm$ 0.09 (0) & 0.43 $\pm$ 0.33 (0) & 0.72 $\pm$ 0.52 (1) \\
MTD: D3 - D & 15 (109) & 5.06 $\pm$ 3.53 (5) & 1.30 $\pm$ 0.90 (1) & 0.10 $\pm$ 0.07 (0) & 0.02 $\pm$ 0.01 (0) & 0.05 $\pm$ 0.03 (0) \\
MTD: D4 - L & 9 (60) & 5.90 $\pm$ 3.55 (6) & 0.35 $\pm$ 0.16 (0) & 0.08 $\pm$ 0.06 (0) & 0.46 $\pm$ 0.32 (0) & 0.14 $\pm$ 0.10 (0) \\
MTD: D4 - D & 10 (90) & 5.06 $\pm$ 5.41 (5) & 0.81 $\pm$ 0.86 (1) & 0.23 $\pm$ 0.25 (0) & 0.18 $\pm$ 0.20 (0) & 0.19 $\pm$ 0.20 (0) \\
MTD: D5 - L & 5 (41) & 14.98 $\pm$ 8.52 (15) & 2.11 $\pm$ 1.21 (2) & 0.19 $\pm$ 0.11 (0) & 0.44 $\pm$ 0.26 (0) & 0.65 $\pm$ 0.36 (1) \\ \hline
\textit{        MTD subtotal} & \textit{93 (665)} & \textit{18.44 $\pm$ 23.82 (18)} & \textit{1.72 $\pm$ 2.17 (2)} & \textit{0.13 $\pm$ 0.13 (0)} & \textit{0.78 $\pm$ 1.14 (1)} & \textit{1.14 $\pm$ 1.70 (1)} \\ \hline
\textbf{TOTAL} & \textbf{104 (1040)} & \textbf{16.85 $\pm$ 16.83 (17)} & \textbf{2.01 $\pm$ 2.09 (2)} & \textbf{0.31 $\pm$ 0.36 (0)} & \textbf{0.82 $\pm$ 0.90 (1)} & \textbf{1.12 $\pm$ 1.20 (1)} \\ \hline
\multicolumn{6}{l}{\begin{minipage}[t]{0.8\textwidth}There were no samples for Day 5 Dinner and no recipes available for foods imaged on Day 2 Lunch. RTF: Regular Texture Foods Dataset; MTD: Modified Texture Foods Dataset; B: Breakfast, L: Lunch, D: Dinner; D\#: Day number. For example, MTD: D1 - L is Modified Texture Foods Dataset: Day 1 - Lunch.\end{minipage}}
\end{tabular}
\end{table}

\begin{table}[]
\caption[Micronutrient intake accuracies of elements within and across datasets.]{Micronutrient intake accuracies of elements within and across datasets. }
\label{tab:nut_est_accuracy_elements}
\small
\centering
\begin{tabular}{llcccc}
\hline
\multicolumn{2}{l}{\textbf{Dataset}} & \multicolumn{4}{c}{\textbf{\begin{tabular}[c]{@{}c@{}}Nutrient intake   accuracy\\      $\mu \pm \sigma$ (\% error)\end{tabular}}} \\
\textbf{Meal} & \textbf{\begin{tabular}[c]{@{}l@{}}\# of classes\\ (\# images)\end{tabular}} & \textbf{Calcium (mg)} & \textbf{Iron (mg)} & \textbf{Sodium (mg)} & \textbf{Zinc (mg)} \\ \hline
RTD: B & 3 (125) & 10.66 $\pm$ 10.09 (11) & 0.26 $\pm$ 0.24 (0) & 20.77 $\pm$ 18.60 (21) & 0.04 $\pm$ 0.03 (0) \\
RTD: L & 3 (125) & 11.96 $\pm$ 12.34 (12) & 0.11 $\pm$ 0.07 (0) & 32.20 $\pm$ 34.33 (32) & 0.00 $\pm$ 0.00 (0) \\
RTD: D & 3 (125) & 2.95 $\pm$ 3.72 (3) & 0.08 $\pm$ 0.09 (0) & 45.71 $\pm$ 46.31 (46) & 0.00 $\pm$ 0.00 (0) \\\hline
\textit{        RTF subtotal} & \textit{9 (375)} & \textit{9.94 $\pm$ 8.48 (10)} & \textit{0.17 $\pm$ 0.18 (0)} & \textit{18.50 $\pm$ 15.58 (18)} & \textit{0.05 $\pm$ 0.06 (0)} \\ \hline
MTD: D1 - L & 15 (90) & 8.45 $\pm$ 14.01 (8) & 0.05 $\pm$ 0.05 (0) & 16.43 $\pm$ 20.69 (16) & 0.02 $\pm$ 0.02 (0) \\
MTD: D1 - D & 5 (25) & 1.86 $\pm$ 1.64 (2) & 0.62 $\pm$ 0.57 (1) & 17.05 $\pm$ 15.11 (17) & 0.50 $\pm$ 0.46 (1) \\
MTD: D2- L & 12 (74) & n/a & n/a & n/a & n/a \\
MTD: D2 - D & 12 (91) & 6.04 $\pm$ 7.03 (6) & 0.17 $\pm$ 0.19 (0) & 11.16 $\pm$ 9.22 (11) & 0.10 $\pm$ 0.12 (0) \\
MTD: D3 - L & 10 (85) & 2.53 $\pm$ 2.03 (3) & 0.09 $\pm$ 0.08 (0) & 13.87 $\pm$ 11.42 (14) & 0.05 $\pm$ 0.04 (0) \\
MTD: D3 - D & 15 (109) & 1.08 $\pm$ 0.75 (1) & 0.02 $\pm$ 0.01 (0) & 0.08 $\pm$ 0.06 (0) & 0.01 $\pm$ 0.01 (0) \\
MTD: D4 - L & 9 (60) & 1.34 $\pm$ 0.93 (1) & 0.03 $\pm$ 0.02 (0) & 12.11 $\pm$ 8.05 (12) & 0.01 $\pm$ 0.01 (0) \\
MTD: D4 - D & 10 (90) & 2.98 $\pm$ 3.22 (3) & 0.07 $\pm$ 0.08 (0) & 3.84 $\pm$ 3.32 (4) & 0.04 $\pm$ 0.04 (0) \\
MTD: D5 - L & 5 (41) & 7.12 $\pm$ 4.54 (7) & 0.12 $\pm$ 0.06 (0) & 18.93 $\pm$ 10.33 (19) & 0.10 $\pm$ 0.06 (0) \\ \hline
\textit{        MTD subtotal} & \textit{93 (665)} & \textit{4.37 $\pm$ 6.19 (4)} & \textit{0.19 $\pm$ 0.34 (0)} & \textit{12.92 $\pm$ 12.73 (13)} & \textit{0.14 $\pm$ 0.27 (0)} \\ \hline
\textbf{TOTAL} & \textbf{104 (1040)} & \textbf{8.51 $\pm$ 8.31 (9)} & \textbf{0.18 $\pm$ 0.23 (0)} & \textbf{17.07 $\pm$ 15.09 (17)} & \textbf{0.07 $\pm$ 0.15 (0)} \\ \hline
\multicolumn{6}{l}{\begin{minipage}[t]{0.8\textwidth}There were no samples for Day 5 Dinner and no recipes available for foods imaged on Day 2 Lunch. RTD: Regular texture foods dataset; MTD: Modified texture foods dataset; B: Breakfast, L: Lunch, D: Dinner; D\#: Day number. For example, MTD: D1 - L is Modified Texture Foods Dataset: Day 1 - Lunch.\end{minipage}}
\end{tabular}
\end{table}

\begin{table}[]
\caption[Micronutrient intake accuracies of vitamins within and across datasets.]{Micronutrient intake accuracies of vitamins within and across datasets.}
\label{tab:nut_est_accuracy_vitamins}
\small
\centering
\begin{tabular}{llcccc}
\hline
\multicolumn{2}{l}{\textbf{Dataset}} & \multicolumn{4}{c}{\textbf{\begin{tabular}[c]{@{}c@{}}Nutrient intake   accuracy\\      $\mu \pm \sigma$ (\% error)\end{tabular}}} \\
\textbf{Meal} & \textbf{\begin{tabular}[c]{@{}l@{}}\# of classes\\ (\# images)\end{tabular}} & \textbf{Vitamin B6 (mg)} & \textbf{Vitamin C (mg)} & \textbf{Vitamin D (IU)} & \textbf{Vitamin K (mcg)} \\ \hline
RTD: B & 3 (125) & 0.02 $\pm$ 0.02 (0) & 0.00 $\pm$ 0.00 (0) & n/s & n/s \\
RTD: L & 3 (125) & 0.00 $\pm$ 0.00 (0) & 5.70 $\pm$ 4.05 (6) & n/s & n/s \\
RTD: D & 3 (125) & 0.00 $\pm$ 0.00 (0) & 0.78 $\pm$ 0.96 (1) & n/s & n/s \\ \hline
\textit{        RTF subtotal} & \textit{9 (375)} & \textit{0.01 $\pm$ 0.02 (0)} & \textit{0.00 $\pm$ 0.00 (0)} & \textit{n/s} & \textit{n/s} \\ \hline
MTD: D1 - L & 15 (90) & 0.01 $\pm$ 0.01 (0) & 1.87 $\pm$ 1.59 (2) & 0.00 $\pm$ 0.01 (0) & 2.02 $\pm$ 2.29 (2) \\
MTD: D1 - D & 5 (25) & 0.10 $\pm$ 0.09 (0) & 0.71 $\pm$ 0.50 (1) & 4.38 $\pm$ 4.03 (4) & 0.32 $\pm$ 0.29 (0) \\
MTD: D2- L & 12 (74) & n/a & n/a & n/a & n/a \\
MTD: D2 - D & 12 (91) & 0.02 $\pm$ 0.03 (0) & 0.18 $\pm$ 0.19 (0) & 0.06 $\pm$ 0.09 (0) & 0.01 $\pm$ 0.01 (0) \\
MTD: D3 - L & 10 (85) & 0.01 $\pm$ 0.00 (0) & 0.31 $\pm$ 0.21 (0) & 0.00 $\pm$ 0.00 (0) & 1.69 $\pm$ 1.18 (2) \\
MTD: D3 - D & 15 (109) & 0.01 $\pm$ 0.01 (0) & 1.70 $\pm$ 1.18 (2) & 0.00 $\pm$ 0.00 (0) & 0.16 $\pm$ 0.11 (0) \\
MTD: D4 - L & 9 (60) & 0.00 $\pm$ 0.00 (0) & 0.66 $\pm$ 0.46 (1) & 0.00 $\pm$ 0.00 (0) & 2.33 $\pm$ 1.63 (2) \\
MTD: D4 - D & 10 (90) & 0.01 $\pm$ 0.01 (0) & 2.80 $\pm$ 3.04 (3) & 0.03 $\pm$ 0.03 (0) & 3.55 $\pm$ 3.86 (4) \\
MTD: D5 - L & 5 (41) & 0.01 $\pm$ 0.01 (0) & 0.96 $\pm$ 0.70 (1) & 0.16 $\pm$ 0.10 (0) & 0.66 $\pm$ 0.36 (1) \\ \hline
\textit{        MTD subtotal} & \textit{93 (665)} & \textit{0.03 $\pm$ 0.05 (0)} & \textit{1.05 $\pm$ 1.50 (1)} & \textit{0.87 $\pm$ 2.41 (1)} & \textit{1.13 $\pm$ 1.98 (1)} \\ \hline
\textbf{TOTAL} & \textbf{104 (1040)} & \textbf{0.02 $\pm$ 0.03 (0)} & \textbf{0.27 $\pm$ 0.88 (0)} & \textbf{0.87 $\pm$ 2.41 (1)} & \textbf{1.13 $\pm$ 1.98 (1)} \\ \hline
\multicolumn{6}{l}{\begin{minipage}[t]{0.8\textwidth}There were no samples for Day 5 Dinner and no recipes available for foods imaged on Day 2 Lunch. RTF: Regular Texture Foods Dataset; MTD: Modified Texture Foods Dataset; B: Breakfast, L: Lunch, D: Dinner; D\#: Day number. For example, MTD: D1 - L is Modified Texture Foods Dataset: Day 1 - Lunch.\end{minipage}}

\end{tabular}
\end{table}

\begin{figure*}
	\centering

	\label{fig:calcarbs}
    \begin{subfigure}[b]{\textwidth}
		\centering
		\includegraphics[width=.9\linewidth]{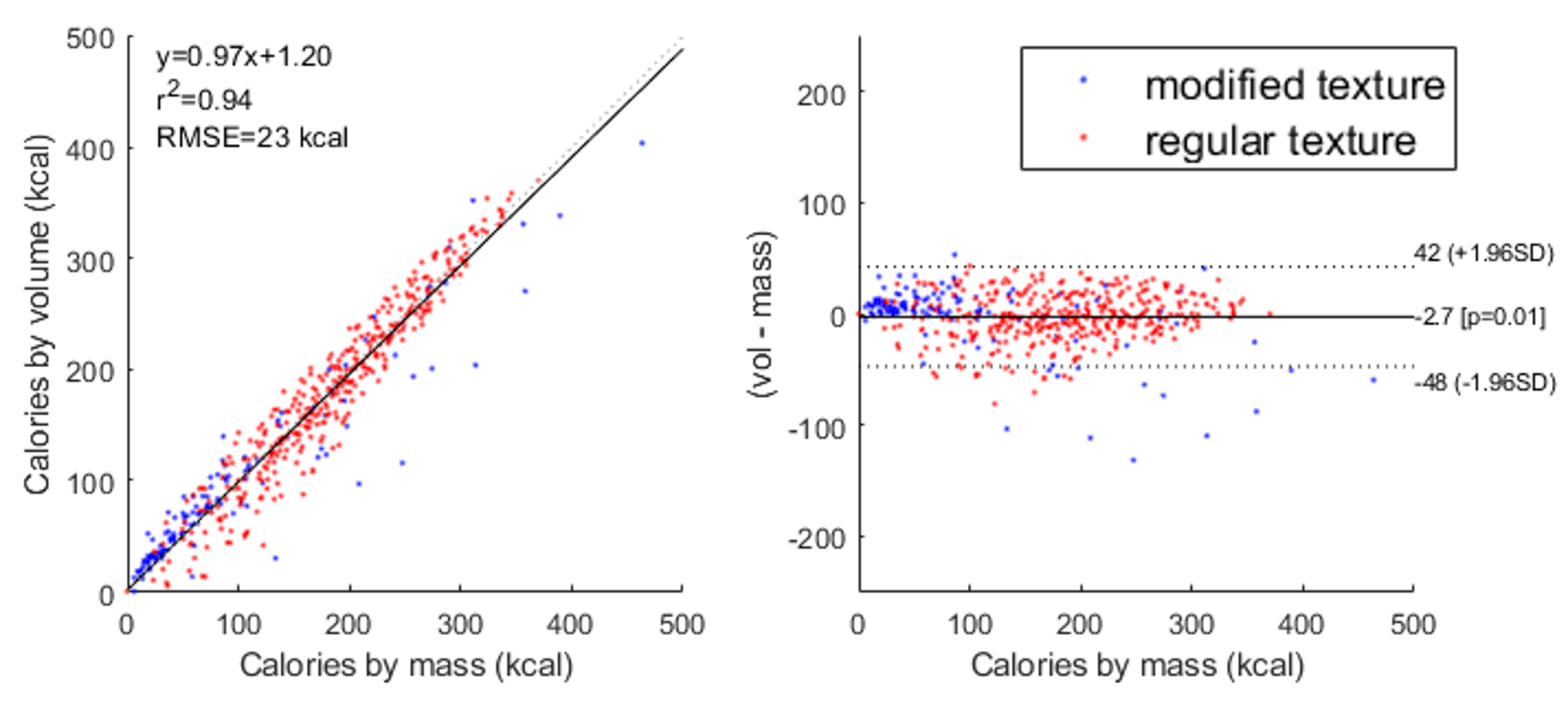}
		\caption[Correlation and agreement between mass and volume estimates calories.]{Correlation and agreement between mass and volume estimates calories.}			
		\label{fig:calories}
	\end{subfigure}%
    
    \begin{subfigure}[b]{\textwidth}
		\centering
		\includegraphics[width=.9\linewidth]{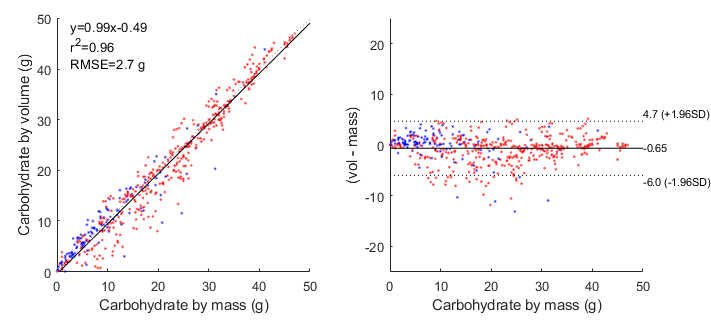}
		\caption[Correlation and agreement between mass and volume estimates carbohydrates.]{Correlation and agreement between mass and volume estimates of carbohydrates.}			
		\label{fig:carbohydrates}
	\end{subfigure}%
    
    \begin{subfigure}[b]{\textwidth}
		\centering
		\includegraphics[width=.9\linewidth]{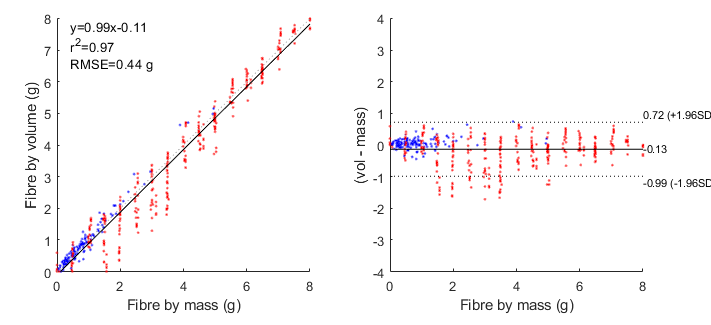}
		\caption{Correlation and agreement between mass and volume estimates of fibre.}			
		\label{fig:fibre}
	\end{subfigure}%
\caption[Nutrients of interest correlation and agreement between mass and volume estimates.]{Correlation and agreement between mass and volume nutrient estimates.}
\end{figure*}

\begin{figure*}
\label{fig:cal_fibre_protein}
	\centering
    \ContinuedFloat
    \begin{subfigure}[b]{\textwidth}
		\centering
		\includegraphics[width=.9\linewidth]{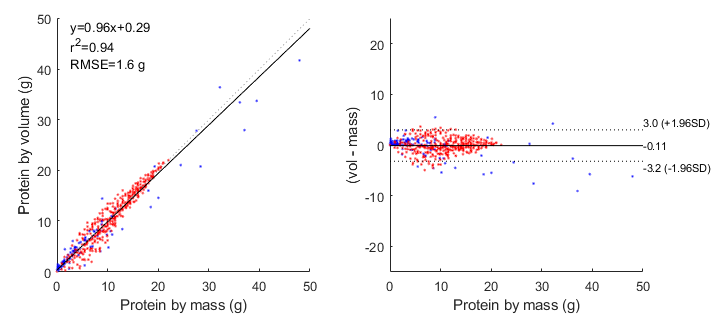}
		\caption{Correlation and agreement between mass and volume estimates of protein.}			
		\label{fig:protein}
	\end{subfigure}%
	
	 \begin{subfigure}[b]{\textwidth}
		\centering
		\includegraphics[width=.9\linewidth]{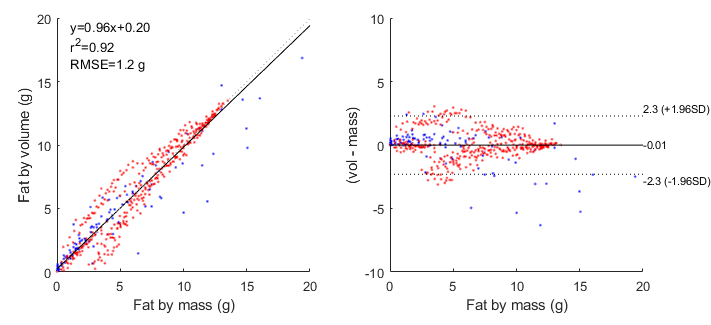}
		\caption{Correlation and agreement between mass and volume estimates of fat.}	\label{fig:fat}
	\end{subfigure}%
    
    \begin{subfigure}[b]{\textwidth}
		\centering
		\includegraphics[width=.9\linewidth]{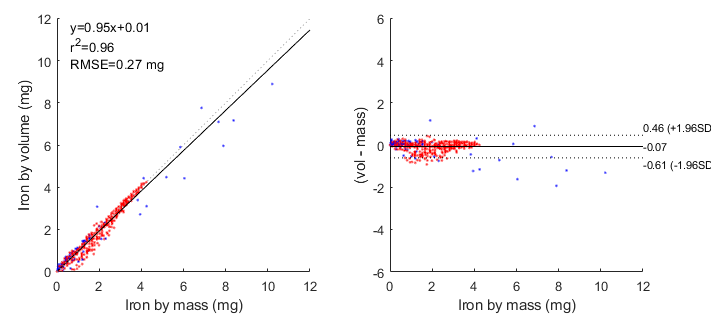}
		\caption{Correlation and agreement between mass and volume estimates of iron.}	\label{fig:iron}
	\end{subfigure}%
	\caption[] {Continued correlation and agreement between mass and volume nutrient estimates.}
\end{figure*}

\begin{figure*}
\label{fig:calcium_sodium}
	\centering
    \ContinuedFloat
    \begin{subfigure}[b]{\textwidth}
		\centering
		\includegraphics[width=.9\linewidth]{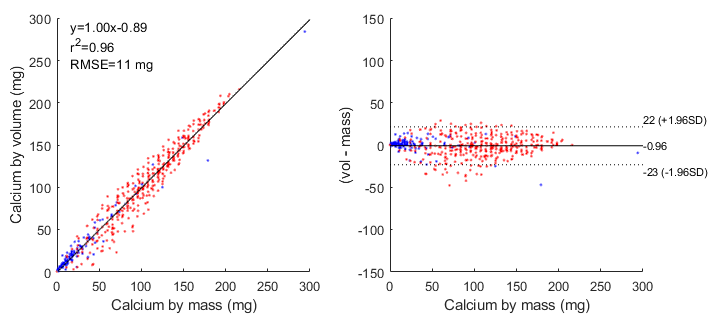}
		\caption{Correlation and agreement between mass and volume estimates of calcium.}			
		\label{fig:calcium}
	\end{subfigure}%
    
    \begin{subfigure}[b]{\textwidth}
		\centering
		\includegraphics[width=.9\linewidth]{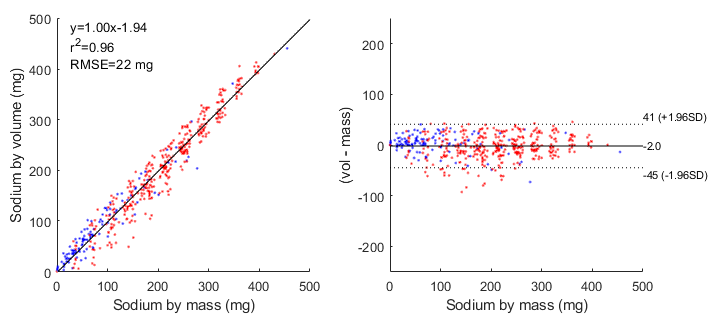}
		\caption{Correlation and agreement between mass and volume estimates of sodium.}	\label{fig:sodium}
	\end{subfigure}%

\begin{subfigure}[b]{\textwidth}
		\centering
		\includegraphics[width=.9\linewidth]{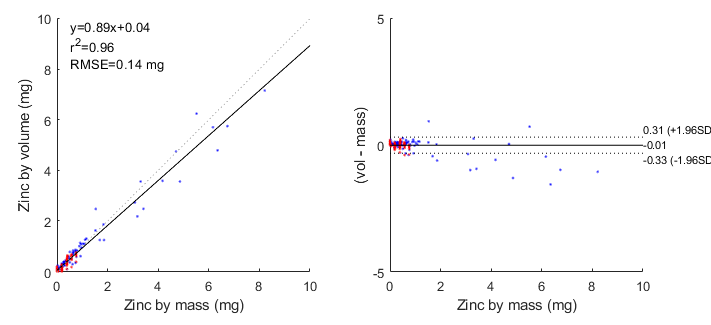}
		\caption{Correlation and agreement between mass and volume estimates of zinc.}	\label{fig:zinc}
	\end{subfigure}%
		\caption[] {Continued correlation and agreement between mass and volume nutrient estimates.}
\end{figure*}

\begin{figure*}
\label{fig:zinc_b6}
	\centering
    \ContinuedFloat

    \begin{subfigure}[b]{\textwidth}
		\centering
		\includegraphics[width=.9\linewidth]{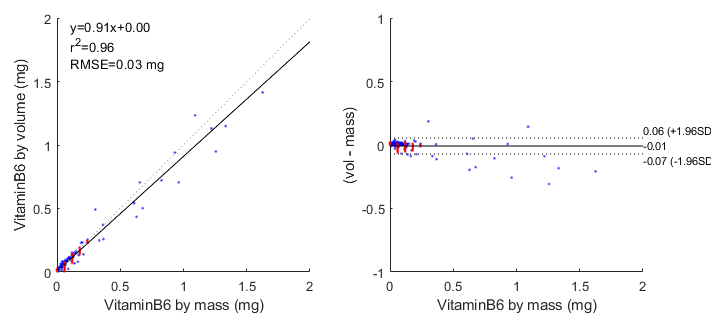}
		\caption{Correlation and agreement between mass and volume estimates of vitamin B6.}			
		\label{fig:vitB6}
	\end{subfigure}%

    \begin{subfigure}[b]{\textwidth}
		\centering
		\includegraphics[width=.9\linewidth]{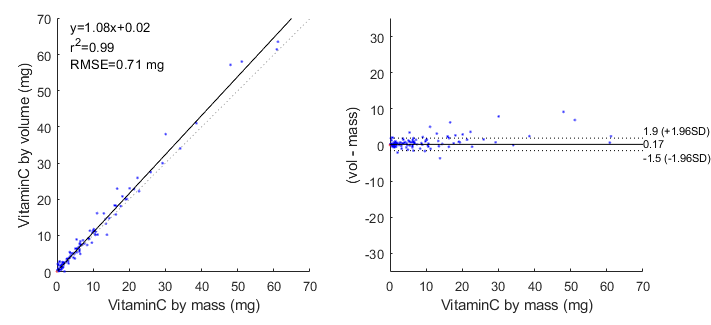}
		\caption{Correlation and agreement between mass and volume estimates of vitamin C.}	\label{fig:VitC}
	\end{subfigure}%
    
    \begin{subfigure}[b]{\textwidth}
		\centering
		\includegraphics[width=.9\linewidth]{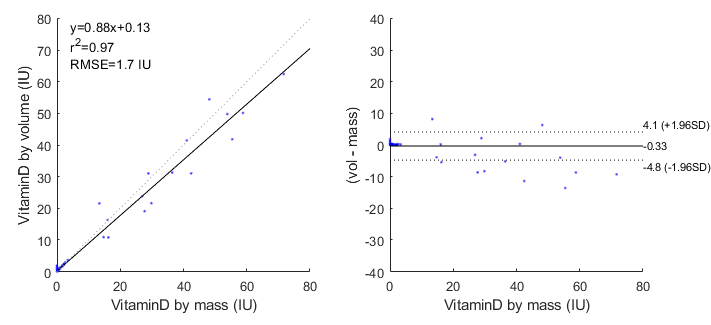}
		\caption{Correlation and agreement between mass and volume estimates of vitamin D.}	\label{fig:vitD}
	\end{subfigure}%
	\caption[] {Continued correlation and agreement between mass and volume nutrient estimates.}
\end{figure*}

\begin{figure*}
\label{fig:K}
	\centering
    \ContinuedFloat 	
    \begin{subfigure}[b]{\textwidth}
		\centering
		\includegraphics[width=.9\linewidth]{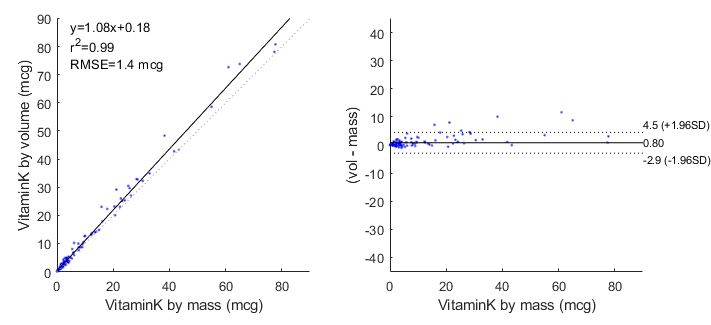}
		\caption{Correlation and agreement between mass and volume estimates of vitamin K.}	\label{fig:vitK}
	\end{subfigure}%
\label{calcarbs}

	\caption[] {Continued correlation and agreement between mass and volume nutrient estimates.}
\end{figure*}

\section*{S3. Standardizing Nutrient Values}
Table~\ref{tab:RDA_conversion} describes the workflow in converting \% daily values to absolute measurements.
\begin{table}[]
\caption[Conversion of \% daily values to absolute.]{Converting the regular texture dataset \% daily values into absolute value to match the modified texture dataset. The Recommended Dietary Allowance (RDA) or Adequate Intake (AI) was used for individuals over 70 years of age and an average across the two sexes was assumed to represent 100\% daily values.}
\small
\centering
\label{tab:RDA_conversion}
\begin{tabular}{lllll}
\hline
 & \multicolumn{3}{c}{\textbf{RDA/AI for > 70 Years}} &  \\ 
\textbf{Nutrient (Input Units)} & \textbf{Males} & \textbf{Females} & \textbf{Assumed 100\% Daily Value} & \textbf{Output Units} \\ \hline
Fat (g) & n/a & n/a & n/a & g \\
Carbohydrates (g) & n/a & n/a & n/a & g \\
Fibre (g) & n/a & n/a & n/a & g \\
Protein (g) & n/a & n/a & n/a & g \\
Calcium\% & 1200 mg & 1200 mg & 1200 mg & mg \\
Iron\% & 8 mg & 8 mg & 8 mg & mg \\
Sodium (mg) & n/a & n/a & n/a & mg \\
Vitamin B6\% & 1.7 mg & 1.5 mg & 1.6 mg & mg \\
Vitamin C\% & 90 mg & 75 mg & 82.5 mg & mg \\
Vitamin D (IU) & ns & ns & ns & IU \\
Vitamin K (mcg) & ns & ns & ns & mcg \\
Zinc (\%) & 11 mg & 8 mg & 9.5 mg & mg \\ \hline
\multicolumn{5}{l}{\begin{minipage}[t]{0.9\textwidth} Where: g: grams, mg: milligrams, IU: international units, mcg: micrograms. While the AI is known for vitamins D and K, no foods in the regular texture foods dataset reported \% daily values and are thus marked as not specified (ns)\end{minipage}}
\end{tabular}
\end{table}

\clearpage

\bibliographystyle{naturemag}

\bibliography{bib}